\def\cvprPaperID{63}
\def\confName{ICCV}
\def\confYear{2023}

\def\paperTitle{YOLOBench: Benchmarking Efficient Object Detectors on Embedded Systems}

\def\authorBlock{
    Ivan Lazarevich \qquad 
    Matteo Grimaldi \qquad 
    Ravish Kumar \qquad
    Saptarshi Mitra \\
    Shahrukh Khan \qquad
    Sudhakar Sah \\
    Deeplite \\
    {\tt\small ivan.lazarevich@deeplite.ai}
}

\newif\ifreview 
\newif\ifarxiv \newcommand{\arxiv}{\arxivtrue}
\newif\ifcamera 
\newif\ifrebuttal 

\arxiv 

\pdfoutput=1
\documentclass[10pt,twocolumn,letterpaper]{article}
\ifreview \usepackage[review]{cvpr} \fi
\ifarxiv \usepackage[pagenumbers]{cvpr} \fi
\ifrebuttal \usepackage[rebuttal]{cvpr} \fi
\ifcamera \usepackage{cvpr} \fi

\usepackage{graphicx}
\usepackage{amsmath}
\usepackage{amssymb}
\usepackage{booktabs}

\usepackage{microtype}
\usepackage{epsfig}
\usepackage[table,xcdraw]{xcolor}
\usepackage{caption}
\usepackage{float}
\usepackage{placeins}
\usepackage{color, colortbl}
\usepackage{stfloats}
\usepackage{enumitem}
\usepackage{tabularx}
\usepackage{xstring}
\usepackage{multirow}
\usepackage{xspace}
\usepackage{url}
\usepackage{subcaption}
\usepackage{xcolor}
\usepackage[hang,flushmargin]{footmisc}

\ifcamera \usepackage[accsupp]{axessibility} \fi

% ...

\ifarxiv  \fi

\newcommand{\R}[1]{{%
    \textbf{%
        \ifstrequal{#1}{1}{\textcolor{red}{R#1}}{%
        \ifstrequal{#1}{2}{\textcolor{blue}{R#1}}{%
        \ifstrequal{#1}{3}{\textcolor{magenta}{R#1}}{%
        \ifstrequal{#1}{4}{\textcolor{teal}{R#1}}{%
                           \textcolor{cyan}{R#1}%
        }}}}%
    }%
}}

\usepackage{xr-hyper}

\makeatletter
\newcommand*{\addFileDependency}[1]{
  \typeout{(#1)}
  \@addtofilelist{#1}
  \IfFileExists{#1}{}{\typeout{No file #1.}}
}

\makeatother

\usepackage[pagebackref,breaklinks,colorlinks]{hyperref}
\usepackage[capitalize]{cleveref}
\crefname{section}{Sec.}{Secs.}
\crefname{table}{Table}{Tables}
\crefname{figure}{Fig.}{Figs.}

\begin{document}
\title{\paperTitle}
\author{\authorBlock}
\maketitle

\begin{abstract}

We present \textit{YOLOBench}, a benchmark comprised of 550+ YOLO-based object detection models on 4 different datasets and 4 different embedded hardware platforms (x86 CPU, ARM CPU, Nvidia GPU, NPU). We collect accuracy and latency numbers for a variety of YOLO-based one-stage detectors at different model scales by performing a fair, controlled comparison of these detectors with a fixed training environment (code and training hyperparameters). Pareto-optimality analysis of the collected data reveals that, if modern detection heads and training techniques are incorporated into the learning process, multiple architectures of the YOLO series achieve a good accuracy-latency trade-off, including older models like YOLOv3 and YOLOv4. We also evaluate training-free accuracy estimators used in neural architecture search on YOLOBench and demonstrate that, while most state-of-the-art zero-cost accuracy estimators are outperformed by a simple baseline like MAC count, some of them can be effectively used to predict Pareto-optimal detection models. We showcase that by using a zero-cost proxy to identify a YOLO architecture competitive against a state-of-the-art YOLOv8 model on a Raspberry Pi 4 CPU. The code and data are available at \url{https://github.com/Deeplite/deeplite-torch-zoo}.

\end{abstract}
\section{Introduction}
\label{sec:intro}

Object detection constitutes a pivotal task in the field of computer vision, entailing the critical process of identifying and localizing objects present within an image. Applications of object detection models include autonomous vehicles, surveillance, robotics, and augmented reality \cite{yolo_applications}. The central problem of deploying deep learning-based object detection solutions on embedded hardware platforms is the amount of computation, memory, and power required for their inference \cite{lane2017squeezing}. This necessitates the development of efficient object detection models specialized for low-footprint hardware devices to achieve an optimal trade-off of accuracy and latency.
\begin{figure*}[ht!]
    \centering
    \includegraphics[width=0.95\textwidth]{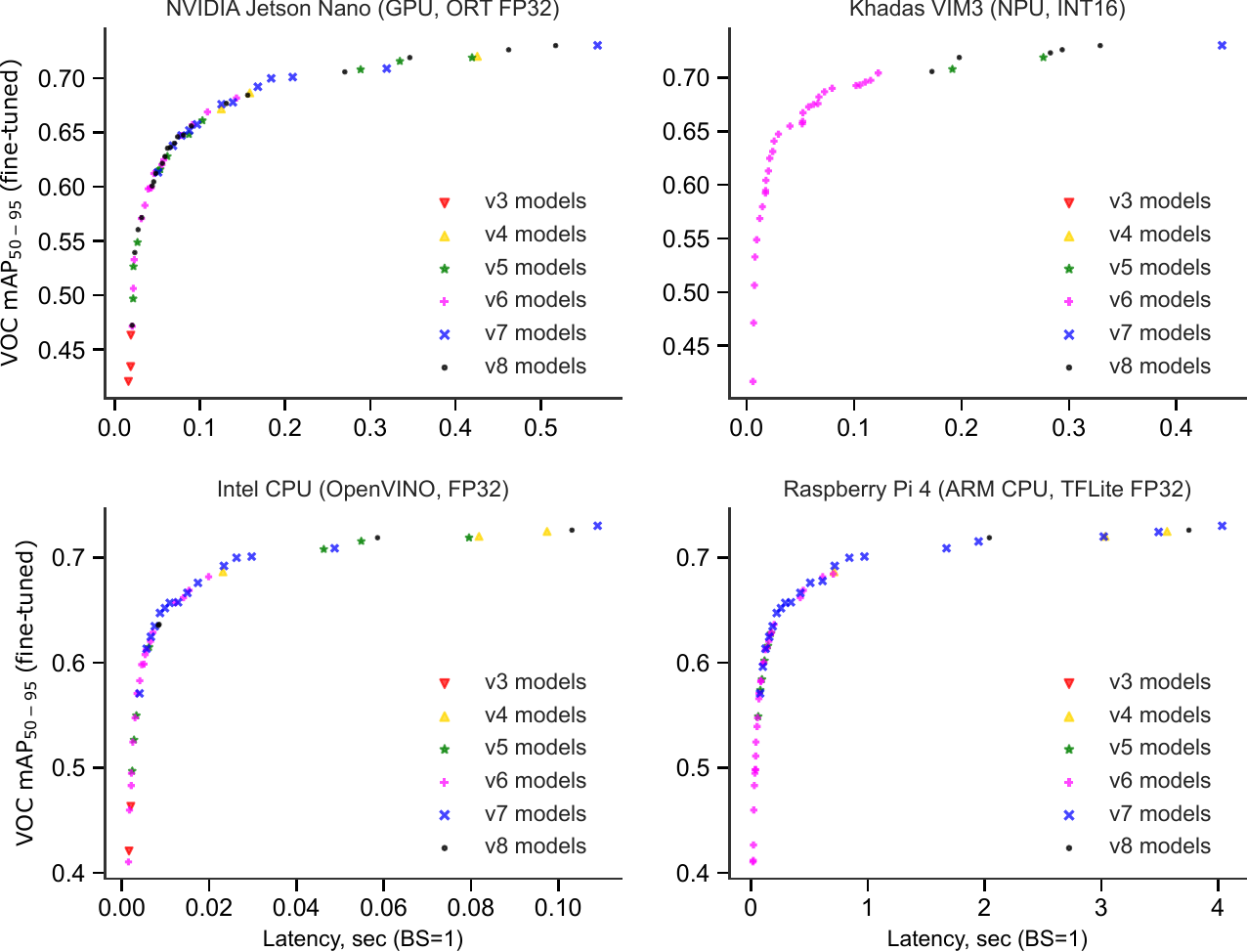}
    \vspace{2mm}
    \caption{Pareto frontiers of \textit{YOLOBench} models fine-tuned on the VOC dataset (on several target resolutions) from COCO-pretrained weights on 4 different hardware platforms. Each point represents a single model in the mAP-latency space, with the model family coded with color and marker shape (all YOLOv6-3.0 models are represented by the same color). Refer to Appendix B for Pareto frontier plots on other datasets.}
    \label{fig:voc_pareto}
\end{figure*}

For years, the state-of-the-art (SOTA) deep learning approach to object detection has been the series of YOLO architectures \cite{yolo_survey}. In recent years, remarkable strides have been taken in advancing YOLO-like single-stage object detectors, prioritizing real-time operation while simultaneously striving for higher accuracy and deployability on low-power devices. These advancements have primarily focused on enhancing various components of the detection pipeline. Key areas of improvement include the design of accurate and efficient backbone and neck structures within the network \cite{wang2023yolov7}, exploration of different detection head designs (e.g. anchor-based \cite{wang2023yolov7} vs. anchor-free \cite{ge2021yolox}), utilization of diverse loss functions \cite{dfl}, and implementation of novel training procedures including innovative data augmentation techniques \cite{dataaug}. 
These collective efforts have continually refined and evolved YOLO-like architectures, enhancing object detection effectiveness and efficiency in real-time scenarios. The differences between consecutive YOLO versions, such as YOLOv5 \cite{Jocher_YOLOv5_by_Ultralytics_2020} and YOLOv6 \cite{li2023yolov6}, span various pipeline components, making it challenging to isolate their individual contributions. This paper aims to address these challenges by providing a fair comparison of recent YOLO versions under controlled conditions (e.g. same training loop for all models) to
demonstrate the impact of the backbone and neck structure of
YOLO-based models in embedded inference applications. 
We also use the collected accuracy and latency data for multiple YOLO-based detector variations to empirically evaluate training-free performance predictors commonly used in neural architecture search \cite{abdelfattah2021zero}. We summarize our contributions as follows: 


\begin{itemize}
    \item We provide a latency-accuracy benchmark of $550$+ YOLO-based object detection models on $4$ different datasets, called \textit{YOLOBench}. All the models are validated on $4$ different embedded hardware platforms (Intel CPU, ARM CPU, Nvidia GPU, NPU),

    \item We show that if modern detection heads and training techniques are implemented for the detector training pipeline, multiple backbone and neck variations, including those of older architectures (e.g. YOLOv3 and YOLOv4), can be used to achieve state-of-the-art latency-accuracy trade-off,
    
    

    \item Looking at \textit{YOLOBench} as a neural architecture search (NAS) space, we demonstrate that, while most of the state-of-the-art zero-cost (training-free) proxies for model accuracy estimation are outperformed by simple baselines such as MAC count, the NWOT estimator \cite{mellor2021neural} can be effectively used to identify potential Pareto-optimal YOLO detectors in a training-free manner,
    

    \item We showcase the effectiveness of the NWOT estimator for
    optimal detector prediction by using it to identify a YOLO-like model with FBNetV3 backbone that outperforms YOLOv8 on the Raspberry Pi 4 ARM CPU.
    
\end{itemize}


\section{Related Work}
\label{sec:related}

\begin{table*}
\small
\caption{Pareto-optimal \textit{YOLOBench} models on 3 datasets and 3 hardware platforms. Shown are the best models in terms of mAP$_{50-95}$ under a given latency threshold (max. latency). For each model, the scaling parameters are given (d33w25 means depth factor $=0.33$ and width factor $=0.25$), corresponding input resolution of the models is indicated in brackets.}
\vspace{1mm}
\label{tab:pareto_table} 
\begin{tabularx}{\linewidth}{lXXXXXX}
\toprule
{HW/max.} & {VOC} & {VOC} & {SKU-110k} & {SKU-110k} & {WIDERFACE} & {WIDERFACE} \\ 
{latency} & {model} & {mAP$_{50-95}$} & {model} & {mAP$_{50-95}$} & {model} & {mAP$_{50-95}$} \\ 
\midrule
{Nano/0.1 sec} & {YOLOv7} & {0.657} & {YOLOv8} & {0.567} & {YOLOv7} & {0.336}\\
{} & {d1w5 (288)} & {} & {d1w25 (480)} & {} & {d1w25 (480)}\\
\midrule
{VIM3/0.05 sec} & {YOLOv6l} & {0.620} & {YOLOv6s} & {0.556} & {YOLOv6m} & {0.318}\\
{} & {d67w25 (416)} & {} & {d33w25 (480)} & {} & {d67w25 (480)}\\
\midrule
{Raspi4/0.5 sec} & {YOLOv6l} & {0.669} & {YOLOv4} & {0.569} & {YOLOv7} & {0.336}\\
{} & {d67w5 (384)} & {} & {d1w25 (480)} & {} & {d1w25 (480)}\\
\bottomrule
\end{tabularx}
\end{table*}

\begin{figure*}[ht!]
    \centering
    \includegraphics[width=0.9\textwidth]{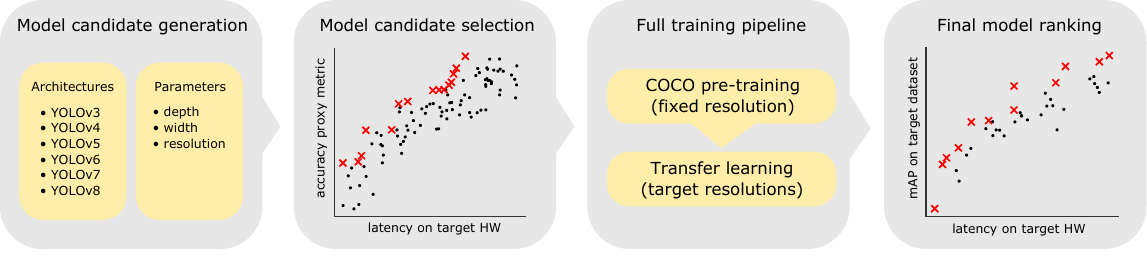}
    \caption{Flowchart of the \textit{YOLOBench} process for model candidate generation, pre-selection and ranking. Pareto-optimal points are depicted as red crosses.}
    \label{fig:scheme}
\end{figure*}

There has been a tremendous amount of progress in efficient object detection in recent years pushing the accuracy-latency frontier, including architectures like YOLOv7 \cite{wang2023yolov7}, YOLOv6-3.0 \cite{li2023yolov6}, DAMO-YOLO \cite{xu2022damo}, RTMDet \cite{lyu2022rtmdet}, RT-DETR \cite{rtdetr} and PP-YOLOE \cite{xu2022pp}. These works oftentimes improve upon state-of-the-art latency-accuracy trade-offs, providing comparisons of several generations of detectors on the COCO dataset. Benchmarks of different model families are also provided by framework developers, such as MMYOLO \cite{mmyolo2022} and Ultralytics \cite{ultralytics}. Additionally, there exist third-party benchmarks of several architectures from the YOLO series on server-grade and embedded GPUs as well as specialized accelerators \cite{stereo_labs,opencv_yolo_benchmarking,feng2022benchmark,zhu2022performance}. We identify a few limitations of the existing efficient detector benchmarks that have served as motivation for \textit{YOLOBench}:

\begin{itemize}
    \item Comparisons of different YOLO versions are frequently done either by using a proxy metric for the actual latency like MAC count and number of parameters or by reporting latency values on server-grade GPUs, neither of which is directly indicative of latency on embedded devices,
    \item Accuracy metrics are usually reported on the COCO dataset, which could be considered too large-scale with respect to actual practical use cases,
    \item Some architecture parameters (like input resolution) are often considered to be fixed in detector benchmarking, while it is known that they serve as important factors in optimal CNN scaling \cite{tinynet},
    \item Different YOLO variations being compared to one another are typically trained with different training codebases, training techniques (loss functions, data augmentations), and hyperparameter values, making it hard to disentangle the contribution of the training pipeline improvements vs. better architecture design. 
\end{itemize}

To address these issues, we conduct a thorough accuracy and latency benchmarking of state-of-the-art YOLO detector versions in controlled, fixed conditions to study the impact of backbone and neck design proposed by several YOLO model families.
\section{Methodology}
\label{sec:method}

\begin{table}
  \small
    \caption{\textit{YOLOBench} architecture space (variation of backbone/neck, depth, width, and input resolution).}
    \label{tab:yolobench_space}
    \vspace*{1mm}
    \begin{tabularx}{\columnwidth}{c|c|c} 
      \textbf{Model} & \textbf{Backbone} & \textbf{Neck}\\
      \hline
      YOLOv3 \cite{redmon2018yolov3} & DarkNet53 & FPN \\
      YOLOv4 \cite{bochkovskiy2020yolov4} & CSPDNet53 & SPP-PAN \\
      YOLOv5 \cite{Jocher_YOLOv5_by_Ultralytics_2020} & CSPDNet53-C3 & SPPF-PAN-C3 \\
      YOLOv6s-3 \cite{li2023yolov6} & EfficientRep & RepBiFPAN\\
      YOLOv6m-3 \cite{li2023yolov6} & CSPBep (e=2/3) & CSPRepBiFPAN\\
      YOLOv6l-3 \cite{li2023yolov6} & CSPBep (e=1/2) & CSPRepBiFPAN\\
      YOLOv7 \cite{wang2023yolov7} & E-ELAN & SPPF-ELAN-PAN \\
      YOLOv8 \cite{ultralytics}&  CSPDNet53-C2f & SPPF-PAN-C2f \\
      \hline
    \end{tabularx}
    \vspace*{1mm}
    \\
    \centering
    Width factor $\in$ \{0.25, 0.5, 0.75, 1.0\} \\
    Depth factor $\in$ \{0.33, 0.67, 1.0\} \\
    Input resolution $\in$ \{160:480:32\}
\end{table}

The purpose of the current study is to thoroughly study the impact of the backbone and neck and its parameters (width, depth, input resolution) on the 
performance of YOLO detectors in terms of their accuracy and latency. For the rest of the factors influencing the accuracy-latency trade-off, such as choice of the detection head, loss function, training pipeline, and hyperparameters, we aim to have a fixed, controlled setup, so that we can isolate the effect of backbone and neck design on model performance. For this reason, we use the anchor-free decoupled detection head of YOLOv8 \cite{yolo_survey}, as well as CIoU and DFL losses for bounding box prediction used in YOLOv8, as they have been shown to produce state-of-the-art results on the COCO dataset. Anchor-free detection in YOLO models has been also shown to provide latency benefits in the end-to-end detection pipelines \cite{rtdetr}. Hence, the main source of variation in \textit{YOLOBench} models is the structure and parameters of the backbone and neck. We also use the same training code and hyperparameters for all models, as set by default in the YOLOv8 training code released by Ultralytics \cite{ultralytics}, which provides a relatively simple training loop capable of producing SOTA results.

The flow of candidate model generation, pre-selection, and training is shown in Figure \ref{fig:scheme}. First, we generate the full architecture space consisting of about $1000$ models by independently varying the backbone/neck structure, depth factor, width factor, and input resolution (Table \ref{tab:yolobench_space}). For each architecture, we consider its variations trained and tested on $11$ different input resolutions (from $160$x$160$ to $480$x$480$ with a step of $32$) and $12$ variations in depth and width, aside from $4$ usually considered scaling variants ($n$, $s$, $m$ and $l$). The only exception is the YOLOv7 models, for which we only vary the width factor producing $4$ variations of the model. For YOLOv6 models, we use the v3.0 version \cite{li2023yolov6}, for which provided $s$, $m$ and $l$ variations actually represent different architectures aside from different depth and width factors (see Table \ref{tab:yolobench_space}). Hence, we consider YOLOv6s, YOLOv6m and YOLOv6l as different model families and generate the same 12 depth-width combinations for each one.

{\bf Latency measurements.} The actual inference latency for each model might vary significantly depending on the deployment environment and runtime. Therefore, we collect the latency measurements for each of the models by running inference on $4$ different hardware platforms (runtime and inference precisions specified in brackets): 

\begin{itemize}
    \item NVIDIA Jetson Nano GPU (ONNX Runtime, FP32)
    \item Khadas VIM3 NPU (AML NPU SDK, INT16)
    \item Raspberry Pi 4 Model B CPU (TFLite with XNNPACK, FP32)
    \item Intel$^{\circledR}$ Core\texttrademark i7-10875H CPU (OpenVINO, FP32)
\end{itemize}

We did not consider latency measurements for INT8 precision, as depending on the quantization scheme (e.g. per-tensor vs. per-channel) and approach (e.g. post-training quantization vs. quantization-aware training), there can be a varied impact of INT8 quantization on accuracy. Adding INT8 results for both accuracy and latency in \textit{YOLOBench} is a matter of future work. All latency measurements were performed with a batch size of $1$ averaged over $200$ inference cycles (with $5$ warmup steps). We measured the inference time required to execute the YOLO model graph, without taking bounding box post-processing (e.g. non-maximum suppression) into account. Note that for VIM3 NPU measurements, the bounding box decoding post-processing operations (operations after the last convolutional layers of the network) were also skipped due to the limitations of VIM3 SDK.

{\bf Training pipeline.} To obtain the accuracy metric values for the models, we consider the following $4$ datasets: (i) PASCAL VOC (20 object categories) \cite{pascal-voc-2012}, (ii) SKU-110k (1 class, retail item detection) \cite{sku-110k}, (iii) WIDER FACE (1 class, face detection) \cite{widerface}, (iv) COCO (80 object categories) \cite{lin2014microsoft}. Our motivation to include several smaller-scale (with respect to COCO) but challenging datasets stems from the fact that for many practical deployment use cases, the number of object categories to detect and the amount of available data might be limited. The metric of interest for all datasets is mAP$_{50-95}$. For all selected models, the training procedure starts with pretraining on the COCO dataset (for $300$ epochs, with a batch size of $64$ and $640$x$640$ resolution), afterward the best COCO weights are used as initialization for other datasets, on which we perform fine-tuning for $100$ epochs (batch size of $64$) on all $11$ \textit{YOLOBench} resolutions and select the best weights (in terms of mAP$_{50-95}$ value) for each one. For the COCO dataset, we do not perform fine-tuning on target resolutions, rather we evaluate the model trained on $640$x$640$ images on all target resolutions (to mimic the deployment of pre-trained COCO weights). All other training hyperparameters are set as per default values of the Ultralytics YOLOv8 codebase \cite{ultralytics}. Model weights are randomly initialized for all experiments (e.g. no transfer of ImageNet weights for the backbone is performed).

{\bf Candidate model pre-selection.} In order to reduce the number of training runs on the COCO dataset, we filter out some of the least promising model candidates from the \textit{YOLOBench} architecture space as an initial step of our benchmarking procedure. To determine the most promising models in terms of the accuracy-latency trade-off, we compute a proxy metric that is well correlated with the final mAP values of the models fine-tuned on the target datasets. A natural choice for such a metric is model performance when trained on a smaller-scale representative dataset. For this purpose, we use the VOC dataset to train all the model candidates from scratch (random initialization) for $100$ epochs and use the resulting mAP$_{50-95}$ value as a proxy metric to predict performance on all target datasets (with models fine-tuned on these datasets from COCO pre-trained weights). We observe a good correlation of such a training-based accuracy proxy with final metrics on all considered datasets (even on datasets from other domains, like SKU-110k; see Appendix C). We also examine the performance of training-free accuracy estimators for this task and compare it to mAP of VOC training from scratch (see Section \ref{sec:ranking_zc}).

Once we have the accuracy proxy values and latency measurements for all models in the dataset, we determine the models with the best accuracy-latency trade-off (the Pareto frontier models). We use the OApackage software library \cite{eendebak2019oapackage} to determine the Pareto optimal elements in the latency-accuracy space. We define the second Pareto set as the set of models that are Pareto-optimal if the initial Pareto set models are removed (so that the ``second best'' models in terms of latency-accuracy trade-off become the best). Correspondingly, we define the $N$-th Pareto set. 

For our model pre-selection procedure, we consider the models contained in the first and second Pareto fronts (in terms of mAP$_{50-95}$ in VOC training from scratch), with latency for each considered hardware platform separately. We merge all the first and second Pareto sets for each HW platform to form the list of promising architectures to be selected for COCO pre-training. After the COCO pre-training phase is finished for a model, variations of that architecture on multiple resolutions are considered in the benchmark.

\begin{figure*}[ht!]
    \centering
    \includegraphics[width=1.\textwidth]{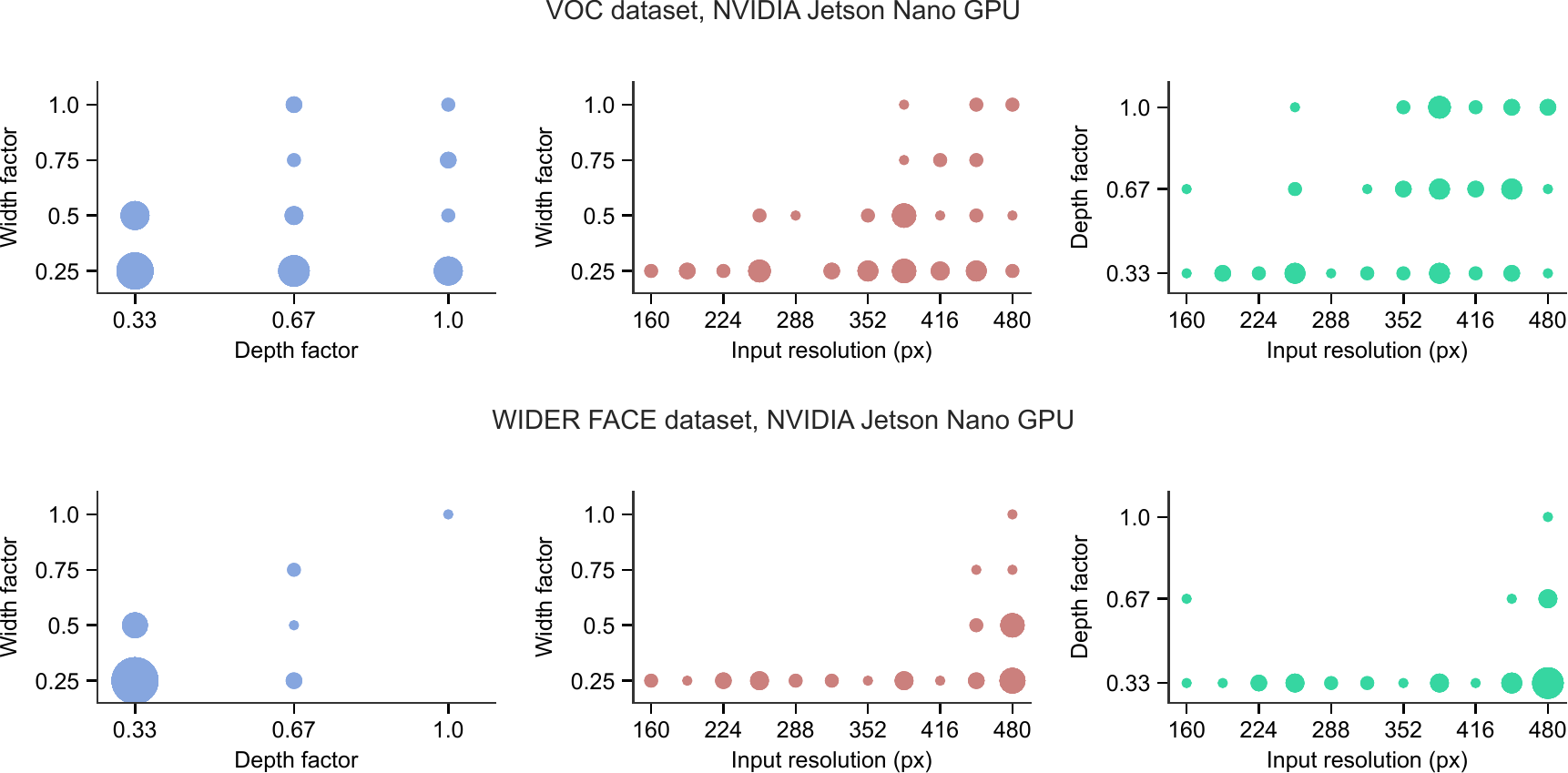}
    \vspace{2mm}
    \caption{Statistics of model scaling parameters (depth factor, width factor, input resolutions) in Pareto-optimal models on VOC and WIDER FACE datasets with latency measured on the Jetson Nano GPU. The size of each point (circle) is proportional to the number of models for that parameter combination.}
    \label{fig:pareto_scaling}
\end{figure*}

\begin{table*}
\small
\caption{Performance of training-free accuracy predictors on \textit{YOLOBench} models and two datasets (VOC and SKU-110k, from COCO-pretrained weights) compared to using metrics of models trained from scratch on the VOC dataset as a predictor. Refer to Appendix C for the data on all considered zero-cost metrics.}
\vspace{2mm}
\label{tab:zc_table}
\begin{tabularx}{\linewidth}{l|XXX|XXX}
\toprule
{} & \multicolumn{3}{c|}{\text{VOC, mAP$_{50-95}$}} & \multicolumn{3}{c}{\text{SKU-110k, mAP$_{50-95}$}}\\
\midrule
{Predictor metric} & {global $\tau$} & {top-15\% $\tau$} & {\%Pareto pred. (GPU)} & {global $\tau$} & {top-15\% $\tau$} & {\%Pareto pred. (GPU)}\\
\midrule
{JacobCov} & {0.095} & {-0.078} & {0.015} & {0.541} & {0.136} & {0.025}\\
{ZiCo} & {0.195} & {0.016} & {0.015} & {0.115} & {0.081} & {0.025}\\
{Zen} & {0.255} & {0.092} & {0.062} & {0.146} & {0.121} & {0.050}\\
{Fisher} & {0.280} & {0.156} & {0.015} & {-0.380} & {-0.096} & {0.025}\\
{SNIP} & {0.336} & {0.217} & {0.015} & {-0.290} & {-0.059} & {0.025}\\
{\#params} & {0.399} & {0.372} & {0.031} & {0.256} & {0.119} & {0.050}\\
{SynFlow} & {0.558} & {0.227} & {0.062} & {0.512} & {0.254} & {0.100}\\
{MACs} & {0.739} & {0.520} & {0.123} & {0.604} & {0.314} & {0.125}\\
{NWOT} & {0.756} & {0.622} & {0.262} & {0.703} & {0.321} & {\bf{0.200}}\\
{NWOT (pre-act)} & {{\bf 0.827}} & {{\bf 0.623}} & {{\bf 0.292}} & {{\bf 0.765}} & {{\bf 0.406}} & {{\bf 0.200}}\\
\midrule
{VOC training} & {0.847} & {0.665} & {0.369} & {0.739} & {0.374} & {0.425}\\
{from scratch (mAP$_{50-95}$)} &  {} & {} & {} & {}\\
\bottomrule
\end{tabularx}
\end{table*}
\section{Results}
\label{sec:results}

\subsection{Pareto-optimal YOLO models}
By computing the proxy metric for model accuracy (mAP$_{50-95}$ in VOC training from scratch) and latency values for the whole \textit{YOLOBench} architecture space on several hardware platforms, we determine the Pareto sets containing the most promising models (in terms of latency-accuracy trade-off) for each HW platform. The first and second Pareto sets for each device are merged into a unified list of best architectures, which is comprised of 52 backbone/neck combinations for COCO pre-training. Same architectures with different input resolutions are considered as the same data points in this list since COCO pre-training is regardless done on a fixed resolution of 640x640. The COCO pre-training phase is followed by fine-tuning at 11 different resolutions (from 160x160 to 480x480 with a step of 32) on all downstream datasets (except for COCO), resulting in 572 models total for each dataset.


\begin{figure*}[ht!]
    \centering
    \includegraphics[width=1.\textwidth]{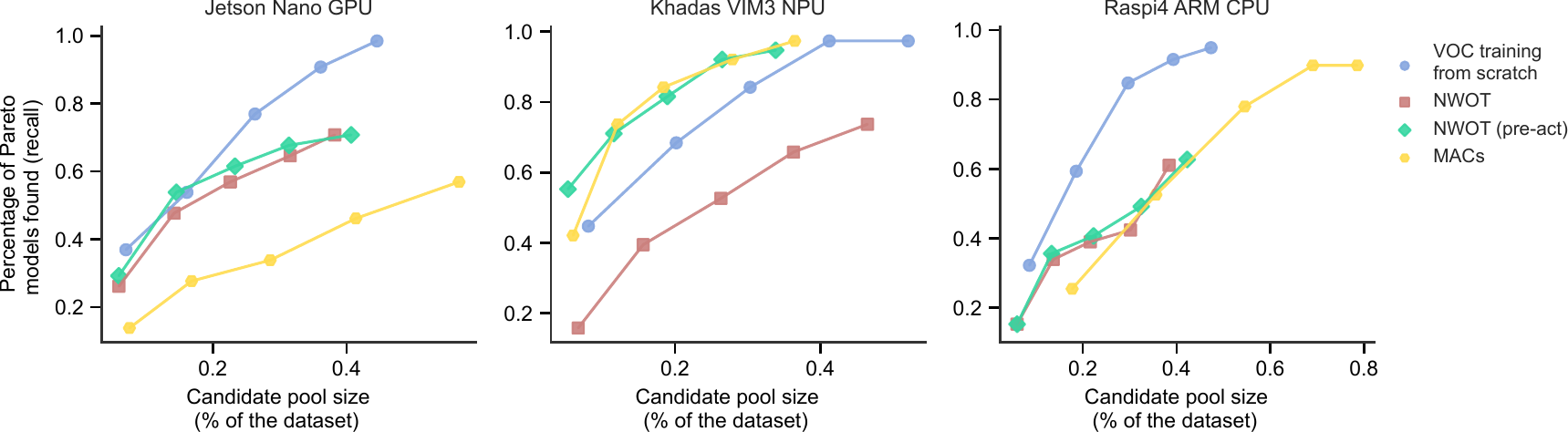}
    \caption{Percentage of all actual Pareto models (recall) found in the candidate pools consisting of first $N$ ($N$ from 1 to 5) ZC-based Pareto sets depending on the HW platform and ZC metric. The data shown is for Pareto-optimal models on the VOC dataset. The x-axis shows the candidate pool size as a percentage of the full dataset size.}
    \label{fig:zc_pareto_voc}
\end{figure*}

Finally, with the obtained fine-tuned model accuracy on several datasets and latency measurements on several devices, we compute the actual Pareto sets for each particular dataset/HW platform combination. Figure \ref{fig:voc_pareto} shows the Pareto frontiers of \textit{YOLOBench} models fine-tuned on the VOC dataset on 4 different devices. Notably, significant differences emerge in these Pareto frontiers between different devices. In particular, the Pareto-optimal set for VIM3 NPU is mostly comprised of YOLOv6 models, with some YOLOv5, YOLOv7, and YOLOv8 models present in the higher accuracy region. This is not found to be the case for the Pareto sets of Intel and ARM CPUs. Despite containing a few YOLOv6 models in the lower latency region, these sets also encompass numerous YOLOv5 and YOLOv7 variations, with limited representation from other model families such as YOLOv3 and YOLOv4.
While the Pareto sets for Intel and ARM CPUs exhibit a certain degree of similarity, the Jetson Nano GPU stands out from the rest of the devices. It showcases a non-uniform distribution of model families, with YOLOv5, YOLOv6, YOLOv7, and YOLOv8 models all represented across the entire accuracy/latency space. Table \ref{tab:pareto_table} shows representative Pareto-optimal models for 3 different datasets (VOC, SKU-110k, WIDERFACE) and 3 hardware platforms under certain latency thresholds. Note that although there are similarities of model family distributions in Pareto sets computed for different datasets (see Appendix B), the exact optimal model for a given latency threshold depends on the specific dataset of interest.

Next, we analyze the statistics of Pareto-optimal models depending on the dataset and hardware platform. Figure \ref{fig:pareto_scaling} shows the distribution of depth factor, width factor, and input resolution values in Pareto frontier models for VOC and SKU-110k datasets on Jetson Nano GPU (data for other datasets and devices are available in Appendix B). The general trend indicates that models at lower input resolutions mostly have lower depth and width factors. This suggests that achieving an optimal latency-accuracy trade-off involves scaling down both the architecture's depth and width before reducing the input resolution. This effect is more pronounced in some datasets (SKU-110k and WIDERFACE), where almost all optimal models are either at the maximal resolution we considered (480x480) with variation in width and depth, or at lower resolutions with minimal width and depth factors. This effect is dataset-dependent, as a more relaxed trend is observed for VOC and COCO datasets, where many optimal models with a variation in width and depth factor are found at resolutions lower than 480x480.

To summarize, we demonstrate that with a state-of-the-art training pipeline and detection head structure, YOLO-based models with various backbone/neck combinations could achieve good latency-accuracy trade-offs in various deployment scenarios, including older backbone/neck structures from YOLOv4 and YOLOv3 models. Furthermore, we show that depth/width reduction precedes input resolution down-scaling in optimal YOLO-based detectors.

\subsection{Ranking training-free accuracy predictors}
\label{sec:ranking_zc}


With an increasing number of architecture blocks and hyperparameter combinations, the size of the candidate model space in \textit{YOLOBench} can further grow exponentially. Hence, it is important to develop efficient methods of filtering out bad architecture proposals before running them through the full training pipeline, including pre-training on the COCO dataset. In the field of neural architecture search, recent works have proposed a handful of training-free, {\em zero-cost} (ZC) estimators that have been shown to perform well on various (relatively simple) benchmarks \cite{mellor2021neural, abdelfattah2021zero, li2023zico}.

Zero-cost estimators were originally proposed by Mellor et al. \cite{mellor2021neural}, and later expanded by Abdelfattah et al. \cite{abdelfattah2021zero} as a means to quickly evaluate the performance of an architecture using only a mini-batch of data. These estimators work by extracting statistics obtained from a forward (and/or backward) pass of a few mini-batches of data through the network, hence eliminating the need for full training of the model. Despite the fact that over 20 different zero-cost accuracy estimators have been introduced in recent years, simple baselines like the number of parameters and MAC count are still found to be hard to outperform \cite{li2023zico}.

The vast search space of YOLO-like architectures necessitates the development of effective training-free estimators to filter out bad candidates and reduce the search space. We examine the performance of a representative subset of zero-cost estimators on \textit{YOLOBench}, namely: 
Fisher~\cite{turner2019blockswap}, GradNorm~\cite{abdelfattah2021zero}, GraSP~\cite{wang2020picking}, JacobCov~\cite{abdelfattah2021zero}, Plain~\cite{abdelfattah2021zero}, SNIP~\cite{lee2018snip}, SynFlow~\cite{tanaka2020pruning}, ZiCo~\cite{li2023zico}, Zen-score~\cite{lin2021zen} and NWOT~\cite{mellor2021neural}. The NWOT metric is computed by measuring the Hamming distance between binary codes produced by each layer's activations \cite{mellor2021neural}. Although originally proposed for ReLU-based networks, we observe that it works well in practice for YOLO variations, most of which contain SiLU activations. The NWOT metric can also be computed by taking the signs of each layer's output features before the activation layer to form the binary code. We refer to that version of the NWOT metric as \textit{NWOT (pre-act)} ("pre-activation") and find that its performance might differ significantly from the original NWOT metric, primarily because the binary codes are computed before the normalization layers followed by the activations. We also compare the performance of the zero-cost predictors with simple baselines such as the number of trainable parameters and MAC count, as well as with a training-based proxy that we have used to pre-select models for \textit{YOLOBench} (mAP$_{50-95}$ in training from scratch on the VOC dataset). 

All zero-cost metrics are computed on randomly initialized models with the same loss function as used for training of all \textit{YOLOBench} models and using a single mini-batch of data with a corresponding image resolution (except for ZiCo, which requires two different mini-batches of data~\cite{li2023zico}). We empirically evaluate the considered set of zero-cost proxies on \textit{YOLOBench} using the following metrics: 
\begin{itemize}
    \item Kendall $\tau$ (global): Kendall rank correlation coefficient evaluated on all \textit{YOLOBench} models
    \item Kendall $\tau$ (top-15\%): Kendall rank correlation coefficient evaluated on the top-15\% performing \textit{YOLOBench} models (in terms of mAP$_{50-95}$ value)
    \item Percentage of all actual Pareto-optimal models in the Pareto set determined with the zero-cost estimator in the zero-cost proxy-latency space (recall for Pareto-optimal model prediction using the ZC-based Pareto set)
\end{itemize}
The last metric effectively measures how accurate the computed Pareto set would be if the proxy values are used instead of actual mAP to rank models. It is calculated by determining Pareto fronts for model rankings based on zero-cost proxies (and real latency measurements) and then estimating how many models present in the actual Pareto set are also present in the ZC-based Pareto set. In other words, a recall value of 0.7 would mean that by taking the models from the ZC-based Pareto set as candidates, we find 70\% of all actual Pareto-optimal models in that candidate set. We report values for Pareto fronts computed with latency measurements on the Jetson Nano GPU in Table \ref{tab:zc_table}.

We generally find that all of the zero-cost predictors we consider (except for NWOT) are outperformed by the simple baseline of MAC count both in terms of Kendall-Tau scores as well as in the percentage of predicted Pareto-optimal models (see Table \ref{tab:zc_table}). Furthermore, when compared with using mAP$_{50-95}$ on VOC training from scratch as a predictor, we observe that only NWOT comes close to it in terms of ranking scores. We also find that the pre-activation version of NWOT tends to work better than standard NWOT on \textit{YOLOBench}. For the task of predicting mAP$_{50-95}$ of models fine-tuned on SKU-110k, we notably observe that pre-activation NWOT outperforms VOC training from scratch metric in terms of Kendall-Tau scores (possibly due to domain difference between VOC and SKU-110k datasets), but the VOC-based proxy metric still performs better for Pareto-optimal model prediction on SKU-110k. For the data on the sensitivity of NWOT predictions to hyperparameter values please refer to Appendix C.

\begin{table}
  \small
    \caption{COCO test and minival mAP and inference latency on Raspberry Pi 4 CPU (TFLite, FP32) for YOLOv8s vs. a model identified from the NWOT-latency Pareto frontier. For latency, mean and standard deviation over 5 runs (each run done for 100 iterations) are shown, with 640x640 input resolution. For mAP, the mean and standard deviation over three random seeds are shown.} 
    \label{tab:timm_coco}
    \vspace*{2mm}
    \begin{tabularx}{\columnwidth}{X|X|X|X} 
      \hline
      {Model} & {mAP$^{test}_{50-95}$} & {mAP$^{val}_{50-95}$} & {Latency, ms}\\
      \hline
      {YOLOv8s} & {43.17\% (0.12\%)} & {44.43\% (0.23\%)} & {1476.09 (1.49)} \\
      \hline
      {YOLOv8s (HSwish)} & {42.90\% (0.00\%)} & {44.23\% (0.10\%)} & {1381.62 (7.34)} \\
      \hline
      {YOLO-FBNetV3-D-PAN-C3} & {\textbf{43.87\%} (0.05\%)} & {\textbf{45.30\%} (0.08\%)} & {\textbf{1355.21} (9.93)} \\
      \hline
    \end{tabularx}
\end{table}

In trying to capture all the real Pareto-optimal models using ZC scores, one could expand the ZC-based candidate pool by calculating subsequent Pareto sets (second, third, fourth, and so forth) and incorporating them into the candidate pool. By applying this strategy, it's possible to identify the complete set of actual Pareto-optimal models while examining only a subset of the entire dataset (e.g., the first $N$ ZC-based Pareto fronts). In this context, we compute candidate pools consisting of $N$ ZC Pareto fronts for each ZC metric and look at the percentage of actual Pareto-optimal models found in the pool versus the pool size (as \% of the full dataset size). Looking at the pool size is motivated by the observation that the number of models in ZC-based Pareto fronts can significantly vary depending on the specific ZC metric used.

Figure \ref{fig:zc_pareto_voc} shows the percentage of predicted real Pareto-optimal models on the VOC dataset contained in pools of $N$ first Pareto fronts for 4 different predictors (VOC training from scratch, NWOT, pre-activation NWOT, and MAC count). For ARM and Intel CPUs, we observe a general trend of VOC training from scratch being the best predictor and MAC count being the worst at all points. Interestingly, for Jetson Nano GPU NWOT performs close to VOC training from scratch for $N = 1,2$ but starts to perform worse with more models in the pool. Surprisingly, MAC count and pre-activation NWOT, which are training-free predictors, outperform VOC training from scratch in predicting Pareto-optimal models on VIM3 NPU.

\subsection{Pareto-optimal detector identification using NWOT score}

To demonstrate the potential of using ZC-based Pareto sets in identifying promising detector architectures with good accuracy-latency trade-off, we additionally generate multiple candidate architectures based on CNN backbones provided by the \texttt{timm} library \cite{rw2019timm}. The architectures are generated by using one of the 347 CNN-based backbones available in \texttt{timm} as a feature extractor followed by a modified Path Aggregation Network (PAN) (same structure with C3 blocks as in YOLOv5 is used, with the number of channels corresponding to YOLOv5s, without the SPPF block) and a YOLOv8 detection head, as in all other \textit{YOLOBench} models. 

We compute the pre-activation NWOT scores as well as measure inference latency on Raspberry Pi 4 ARM CPU with TFLite for all candidate models. We then use the NWOT score and latency values for each model to compute the Pareto frontier in the NWOT-latency space (see Appendix D). We then train one of the models identified to belong to the NWOT-based Pareto frontier (YOLO with FBNetV3-D backbone) on the COCO dataset with a similar setup used to pre-train \textit{YOLOBench} models (640x640 input resolution, 500 epochs, batch size 256, other hyperparameters set to default of Ultralytics YOLOv8 \cite{ultralytics})\footnote{Note that YOLOv8s results provided by Ultralytics \cite{ultralytics} are slightly higher than the ones we report. However, no script to reproduce those results has been released to date.}. The resulting model is found to be more accurate and faster than YOLOv8s (a model in a similar latency range) when tested on Raspberry Pi 4 CPU with TFLite (FP32, XNNPACK backend) (see Table \ref{tab:timm_coco}). Furthermore, we look at the accuracy and latency of a YOLOv8s modification with SiLU activations replaced with Hardswish activations (Table \ref{tab:timm_coco}), as we observe the choice of activation function to be a significant factor affecting TFLite inference latency. We find that the identified NWOT-Pareto model (also containing Hardswish activations in the backbone, neck, and head) still outperforms YOLOv8s-HSwish in terms of latency and accuracy.

\section{Conclusion}
\label{sec:conclusion}

In this work, we present \textit{YOLOBench}, a latency-accuracy benchmark of several hundred YOLO-based models on 4 different object detection datasets and 4 different hardware platforms. The accuracy and latency data are collected in a fixed, controlled environment with the only variation in backbone/neck structure and input image resolution of the detectors. We use these data to demonstrate that it is possible to achieve Pareto-optimal results with a range of different backbone structures, including those of the older architectures in the YOLO series, such as YOLOv3 and YOLOv4. We also observe that depth and width scaling precede input resolution scaling in optimal YOLO-based detectors. 

Finally, we use \textit{YOLOBench} to evaluate zero-cost accuracy predictors, and find that, while many of the existing state-of-the-art predictors perform poorly, pre-activation NWOT score can be effectively used to identify Pareto-optimal detectors for specific target devices of interest. We demonstrate that by using NWOT to find a YOLO backbone (FBNetV3-D) that outperforms a state-of-the-art YOLOv8 model when deployed on a Raspberry Pi 4 ARM CPU.

{\small
\bibliographystyle{ieee_fullname}
\bibliography{11_references}
}

\ifarxiv \clearpage \renewcommand{\thefigure}{S\arabic{figure}}
\renewcommand{\thetable}{S\arabic{table}}

\appendix
\label{sec:appendix}

\section{Latency measurements.}
\label{sec:appendix_A}

Details regarding hardware platforms used to collect latency measurements are outlined in Table \ref{tab:benchmarking_hardware}. Figures \ref{fig:nano_timing_corr} and \ref{fig:raspi4_timing_corr} show the difference of latency value distributions between devices computed for the full initial \textit{YOLOBench} architecture space consisting of $\sim$1000 models. While generally good correlation is observed between model inference latencies on different devices (see also Figure \ref{fig:timing_corr_matrix}), notably latency values measured on Khadas VIM3 NPU differ significantly from latency values on other devices. That is, for models with roughly the same latency on Jetson Nano GPU or Raspi4 ARM CPU, the difference in VIM3 NPU latency could be up to several times.  This difference between NPU values from other common GPU/CPU-based platforms highlights the necessity to develop hardware-aware architecture design and search methods. The difference in the NPU benchmark is also reflected in the structure of model Pareto frontiers (Figs. \ref{fig:voc_pareto}, \ref{fig:coco_pareto}, \ref{fig:sku_pareto}) and the performance of zero-cost predictors in identifying Pareto-optimal models (Figs. \ref{fig:zc_pareto_voc}, \ref{fig:zc_sku_parero_front}). 

\section{\textit{YOLOBench} Pareto frontiers for different datasets.}
\label{sec:appendix_B}

\textit{YOLOBench} Pareto frontiers for SKU-110k, WIDER FACE, and COCO datasets are shown in Figs. \ref{fig:sku_pareto}, \ref{fig:wider_pareto}, \ref{fig:coco_pareto}, correspondingly. Note that while mAP$_{50-95}$ values for VOC, SKU-110k, and WIDER FACE datasets are obtained by fine-tuning COCO pre-trained weights (all trained at 640x640 image resolution) on multiple image resolutions considered in \textit{YOLOBench} (11 values from 160 to 480 with a step of 32), the mAP$_{50-95}$ values on the COCO dataset are obtained by directly evaluating pre-trained COCO weights, without fine-tuning on the corresponding target image resolutions. This corresponds to the situation of deployment of pre-trained COCO weights without any additional training.

Table \ref{tab:pareto_table_full} shows the identified Pareto-optimal YOLO models on 3 different datasets and 4 hardware platforms under several latency thresholds. It can be noted that under the same latency threshold on a given hardware platform, the optimal YOLO model family and input image resolution are typically dataset-dependent.

Figures \ref{fig:scaling_raspi4} and \ref{fig:scaling_vim3} show the statistics of architecture scaling parameters (width factor, depth factor, image resolution) in Pareto-optimal models on Raspberry Pi4 CPU and VIM3 NPU, respectively. Although some differences are observed between devices and datasets (in particular depth factor distributions), there is a general trend in all computed Pareto fronts where a variation in depth/width factors is observed at higher resolutions, and resolution is reduced when the depth/width factors (especially the width factor) already have low values.

\begin{table*}[ht!]
\caption{Details on hardware platforms and corresponding runtimes used for benchmarking.}
\vspace*{3mm}
\label{tab:benchmarking_hardware}
\resizebox{\linewidth}{!}{
\begin{tabular}{l|c|c|c|c}
\toprule
{\bf } & {\bf Raspberry Pi 4 Model B} & {\bf Jetson Nano (NVIDIA)} & {\bf Khadas VIM3} & {\bf Lambda tensorbook}\\
\midrule
{CPU} & {\vtop{\hbox{\strut Quad Core Cortex-A72,}\hbox{\strut 64-bit SoC @1.8GHz}}} & {\vtop{\hbox{\strut Quad Core Cortex-A57 MPCore,}\hbox{\strut 64-bit SoC @1.43GHz}}} & {\vtop{\hbox{\strut Quad Core Cortex-A73 @2.2Ghz,}\hbox{\strut Dual Core Cortex-A53 @1.8Ghz}}} & {\vtop{\hbox{\strut Intel$^{\circledR}$ Core\texttrademark i7-10875H CPU}\hbox{\strut @ 2.30GHz}}} \\
\midrule
{Memory} & {4GB LPDDR4-3200 SDRAM} & {\vtop{\hbox{\strut 4 GB 64-bit LPDDR4,}\hbox{\strut 1600MHz 25.6 GB/s}}} & {4GB LPDDR4/4X} & {64GB DDR4 SDRAM}\\
\midrule
{AI-chip} & {-} & {\vtop{\hbox{\strut NVIDIA Maxwell GPU,}\hbox{\strut 128 NVIDIA CUDA$^{\circledR}$ cores}}} & {\vtop{\hbox{\strut Custom NPU}\hbox{\strut INT8 inference up to 1536 MAC}}} & {NVIDIA RTX 2080 Super Max-Q} \\
\midrule
{Ops}  & {-}& {472 GFLOPs} & {5.0 TOPS}& {-}\\
\midrule 
{Framework/runtime} & {TensorFlow Lite (FP32, XNNPACK backend)}& { ONNX Runtime (FP32, GPU)} & {AML NPU SDK (INT16)} & {OpenVINO (CPU, FP32)}\\
\bottomrule
\end{tabular}}
\end{table*}
 
\section{Performance of zero-cost accuracy predictors on \textit{YOLOBench}.}
\label{sec:appendix_C}

The performance of zero-cost accuracy predictors used in neural architecture search \cite{abdelfattah2021zero} is empirically evaluated on \textit{YOLOBench} models on VOC and SKU-110k. Table \ref{tab:zc_table_full} shows the Kendall-Tau scores and Pareto-optimal model prediction recall values obtained by a variety of zero-cost predictors. The zero-cost predictor values are computed using a randomly sampled batch of test set data with batch size $= 16$ (the used batch was the same for all ZC metrics). MAC count and the number of parameters are computed for models in evaluation mode, with normalization layers fused into preceding convolutions (if possible), and RepVGG-style blocks \cite{ding2021repvgg} also fused, if present in the model. Hence, the performance of MAC and parameter counts might slightly differ if computed for models in training mode. Most predictors perform poorly and are outperformed by the MAC count baseline, except for the NWOT score (in particular the pre-activation version of it). The good performance of NWOT can be also observed in Fig. \ref{fig:ZC_scatter}, where scatter plots of fine-tuned model mAP$_{50-95}$ vs. zero-cost predictor value are shown for a few predictors. Some predictors (notably parameter count, ZiCo, and Zen-score) can be observed to produce very close values for subsets of models with significantly different accuracy. This is an indication of the fact that these predictors perform poorly in estimating accuracy differences in models when the underlying architecture is fixed, but the input image resolution is varied. 

We also test the performance of a training-based predictor on \textit{YOLOBench} which is the mAP$_{50-95}$ values of models trained on a representative dataset (VOC) from scratch for 100 epochs. This predictor sets a strong baseline to be outperformed by training-free predictors, as it is generally found to perform well on a variety of datasets (see Fig. \ref{fig:voc_scratch_corr}), including datasets from different visual domains (e.g. SKU-110k).

We further look into the robustness of the results obtained with the pre-activation NWOT estimator. Since this zero-cost estimator does not require computing the loss function, the main parameters that could influence its performance are the exact batch of data sampled, the batch size, and the dataset split (training or test data) used to sample the batch. Figure \ref{fig:nwot_robust} shows the global Kendall-Tau scores achieved with pre-activation NWOT on VOC \textit{YOLOBench} models with different batches sampled, different batch sizes and different data splits used. There is an observed variance in performance depending on the sampled batch, which is higher when the test set data are used (with an absolute difference of up to $0.05$ in global Kendall-Tau score). Notably, scores computed on training set data (with augmentations) performed better on average compared to test set data, and performance is observed to decrease with increasing batch size. Furthermore, Table \ref{tab:nwot_robust_hohead} shows the mean and standard deviation of Kendall-Tau scores for the standard and pre-activation versions of NWOT on VOC \textit{YOLOBench} models computed on 5 different batches of size 16. We also estimate the performance of the mean predictor values averaged over the 5 sampled batches, which is expectedly found to outperform predictors computed on single batches. Moreover, we compute the pre-activation NWOT scores for all layers in YOLO models except the ones contained in detection heads. This is motivated by the fact that the larger distances between binary activation codes in NWOT are meant to correlate with better performance for the feature extraction layers (e.g. layers in the backbone and neck of YOLO), not the last layers used to compute model predictions. We find an overall performance boost in terms of Kendall-Tau scores for the case when the NWOT score is computed only for the layers in the backbone and neck (Table \ref{tab:nwot_robust_hohead}).

\section{Pareto-optimal model prediction using training-free proxies.}
\label{sec:appendix_D}

We evaluate the training-free accuracy predictors (and the training-based one, VOC training from scratch) for the task of predicting Pareto-optimal models. That is, if one computes the ZC values for each model and determines the Pareto set of models in the ZC value-real latency two-dimensional space, we want to estimate how many models in that Pareto set are going to also be present in the actual Pareto frontier (computed in the two-dimensional mAP$_{50-95}$-latency space). Two metrics are of importance here: recall (how many of actual mAP-latency Pareto-optimal models are captured by a ZC-based Pareto set) and precision (how many of ZC-based Pareto set models are actually Pareto optimal in the real mAP-latency space). Additionally, one could consider the first $N$ ($N=1, 2, 3,...$) ZC-based Pareto sets to expand the set of potential model candidates. We look at how precision and recall values change with $N$ for a few well-performing predictors (NWOT, pre-activation NWOT, MAC count, and VOC training from scratch) with latency values taken from different target devices. 

Recall values for several zero-cost predictors for Pareto models on Jetson Nano GPU and VOC dataset are shown in Fig. \ref{fig:ZC_pareto_VOC_nano}. Corresponding precision values for a few well-performing predictors on 3 different HW platforms are shown in Fig. \ref{fig:ZC_pareto_VOC_precision}. Recall values for these best-performing predictors on the SKU-110k dataset are shown in Fig. \ref{fig:zc_sku_parero_front}.

A different way to evaluate the predictors on \textit{YOLOBench} is to treat models with the same architectures but different input image resolutions as identical data points. That is, if a certain architecture is predicted by ZC-based Pareto front to be optimal on a certain resolution, we count that as a correct prediction if that same architecture on a different resolution is found to be really Pareto-optimal. Such a way to evaluate ZC performance stems from the fact that in practice one typically wishes to predict the most promising architectures, not necessarily the particular optimal image resolution (since that architecture would be pre-trained with a certain fixed resolution, e.g. 640x640 on a dataset like COCO for further fine-tuning on the target dataset). Recall and precision values for such an evaluation protocol for the VOC dataset are shown in Figs. \ref{fig:ZC_pareto_VOC_nores}, \ref{fig:ZC_pareto_VOC_nores_precision}.

We also evaluate the performance of the best training-free predictor (pre-activation NWOT) in predicting Pareto-optimal models, when the latency values used are different from actual latency measurements, but either are computed via a latency proxy like MAC count or measurement on another device. Note that in the case of MAC count as a latency predictor, the whole Pareto-frontier computation process is zero-cost: the approximation for mAP is given by the pre-activation NWOT score, the approximation for latency by MAC count. One might wonder how such a fully zero-cost approach performs in practice. Figures \ref{fig:ZC_pareto_VOC_latproxy} and \ref{fig:ZC_pareto_VOC_latproxy_prec} show the recall and precision values when accuracy predictor is taken to be pre-activation NWOT and latency predictors are varied from MAC count to latencies from other (proxy) devices. Interestingly, MAC count is found to perform relatively well in terms of recall, specifically for Raspberry Pi 4 CPU. Notably, none of the latency proxies work well to predict Pareto-optimal models on VIM3 NPU. Also, perhaps not surprisingly, using Intel CPU latency measurements works well to predict Pareto-optimal models on Raspberry Pi 4 CPU, but does not significantly outperform MAC count.

Finally, we test the pre-activation NWOT accuracy estimator to predict potentially well-performing models out of a set of YOLO models we generated with different CNN backbones from the \texttt{timm} package \cite{rw2019timm}. We have computed the NWOT-latency Pareto set for YOLO-PAN-C3 models with \texttt{timm} backbones on input images of 480x480 resolution, with latency measured on Raspberry Pi 4 ARM CPU (TFLite, FP32). The neck structure (PAN-C3) for each of the candidate models was taken to be that of YOLOv5s and the detection head to be that of YOLOv8 (same as for all \textit{YOLOBench} models), with Hardswish activations in the neck and head, and activation function(s) in the backbone kept the same as originally implemented in \texttt{timm}. Table \ref{tab:nwot_timm_pareto} shows examples of predicted Pareto-optimal models (a subset of the full NWOT-latency Pareto set). Based on these observations, we have selected FBNetV3-D as a potential backbone of a YOLO model to be trained on the COCO dataset and compared it to a reference YOLOv8 model in a similar latency range (YOLOv8s). 

Table \ref{tab:timm_coco_full} shows COCO minival mAP$_{50-95}$ and inference latency results for a YOLO-FBNetV3-D-PAN-C3 model trained on the COCO dataset for 300 epochs and profiled on 640x640 input resolution on Raspi4 CPU with TFLite. We observe that the choice of activation function significantly affects TFLite model inference latency, so for a more fair comparison we also train and profile a Hardswish-based version of YOLOv8s in addition to its default SiLU-based version. While we observe a significant reduction in inference latency with a negligible mAP drop shifting from SiLU to Hardswish, the FBNetV3-based model still outperforms YOLOv8s-HSwish. Furthermore, we train and profile a ReLU-based version of YOLO-FBNetV3-D-PAN-C3 (with activation functions in the backbone kept to be those of the original backbone, i.e. Hardswish, but neck and detection head activations replaced with ReLU) and observe further latency improvements at the cost of $\sim 0.56\%$ drop in mAP$_{50-95}$. However, this model is still found to outperform YOLOv8s in terms of both accuracy and latency (see Table \ref{tab:timm_coco_full}). Furthermore, we train the same models for 500 epochs with a batch size of $256$, which is found to achieve better results on COCO minival and test (Table \ref{tab:timm_coco}). Although we could not exactly reproduce COCO minival mAP results for YOLOv8s reported by Ultralytics \cite{ultralytics}, we find that the FBNetV3-based model outperforms both our YOLOv8s mAP results as well as those of Ultralytics, with lower latency on Raspberry Pi 4 CPU. The COCO minival mAP$_{50-95}$ values reported in Table \ref{tab:timm_coco} were obtained using \texttt{pycocotools} \cite{pycocotools} (with IoU threshold for NMS $=0.6$ and object confidence threshold for detection $=0.001$), and mAP values on test-dev were obtained using the same evaluation parameters by submitting to the competition server \cite{coco_test}. More details on the performance comparison of models on COCO test-dev are shown in Table \ref{tab:timm_coco_test}. 


\begin{table*}
\small
\caption{Pareto-optimal \textit{YOLOBench} models on 3 datasets and 4 hardware platforms. Shown are the best models in terms of mAP$_{50-95}$ under a given latency threshold (max. latency). For each model, the scaling parameters are given (d33w25 means depth factor $=0.33$ and width factor $=0.25$), corresponding input resolution of the models is indicated in brackets.}
\vspace{1mm}
\label{tab:pareto_table_full} 
\begin{tabularx}{\linewidth}{lXXXXXX}
\toprule
{HW/max.} & {VOC} & {VOC} & {SKU-110k} & {SKU-110k} & {WIDERFACE} & {WIDERFACE} \\ 
{latency} & {model} & {mAP$_{50-95}$} & {model} & {mAP$_{50-95}$} & {model} & {mAP$_{50-95}$} \\ 
\midrule
{Nano/0.5 sec} & {YOLOv8} & {0.726} & {YOLOv7} & {0.593} & {YOLOv7} & {0.382}\\
{} & {d67w1 (448)} & {} & {d1w75 (480)} & {} & {d1w75 (480)} & {}\\
{Nano/0.3 sec} & {YOLOv7} & {0.701} & {YOLOv7} & {0.589} & {YOLOv7} & {0.369}\\
{} & {d1w5 (480)} & {} & {d1w5 (480)} & {} & {d1w5 (480)}\\
{Nano/0.1 sec} & {YOLOv7} & {0.657} & {YOLOv8} & {0.567} & {YOLOv7} & {0.336}\\
{} & {d1w5 (288)} & {} & {d1w25 (480)} & {} & {d1w25 (480)}\\
\midrule
{VIM3/0.3 sec} & {YOLOv8} & {0.726} & {YOLOv7} & {0.593} & {YOLOv7} & {0.382}\\
{} & {d67w1 (448)} & {} & {d1w75 (480)} & {} & {d1w75 (480)} & {}\\
{VIM3/0.1 sec} & {YOLOv6l} & {0.669} & {YOLOv8} & {0.567} & {YOLOv6m} & {0.350}\\
{} & {d67w5 (384)} & {} & {d1w25 (480)} & {} & {d33w5 (480)}\\
{VIM3/0.05 sec} & {YOLOv6l} & {0.620} & {YOLOv6s} & {0.556} & {YOLOv6m} & {0.318}\\
{} & {d67w25 (416)} & {} & {d33w25 (480)} & {} & {d67w25 (480)}\\
\midrule
{Intel/0.08 sec} & {YOLOv8} & {0.719} & {YOLOv7} & {0.593} & {YOLOv7} & {0.382}\\
{} & {d1w75 (416)} & {} & {d1w75 (480)} & {} & {d1w75 (480)} & {}\\
{Intel/0.04 sec} & {YOLOv7} & {0.701} & {YOLOv7} & {0.589} & {YOLOv7} & {0.369}\\
{} & {d1w5 (480)} & {} & {d1w5 (480)} & {} & {d1w5 (480)}\\
{Intel/0.02 sec} & {YOLOv6l} & {0.682} & {YOLOv6l} & {0.576} & {YOLOv6l} & {0.346}\\
{} & {d6w5 (448)} & {} & {d33w5 (480)} & {} & {d33w5 (480)}\\
\midrule
{Raspi4/3 sec} & {YOLOv8} & {0.719} & {YOLOv7} & {0.593} & {YOLOv7} & {0.382}\\
{} & {d1w75 (416)} & {} & {d1w75 (480)} & {} & {d1w75 (480)} & {}\\
{Raspi4/1 sec} & {YOLOv7} & {0.701} & {YOLOv7} & {0.589} & {YOLOv7} & {  0.369}\\
{} & {d1w5 (480)} & {} & {d1w5 (480)} & {} & {d1w5 (480)}\\
{Raspi4/0.5 sec} & {YOLOv6l} & {0.669} & {YOLOv4} & {0.569} & {YOLOv7} & {0.336}\\
{} & {d67w5 (384)} & {} & {d1w25 (480)} & {} & {d1w25 (480)}\\
\bottomrule
\end{tabularx}
\end{table*}

\begin{table*}
\small
\caption{Performance of training-free accuracy predictors on \textit{YOLOBench} models and two datasets (VOC and SKU-110k, from COCO-pretrained weights) compared to using mAP$_{50-95}$ of models trained from scratch on the VOC dataset as a predictor.}
\vspace{2mm}
\label{tab:zc_table_full}
\begin{tabularx}{\linewidth}{l|XXX|XXX}
\toprule
{} & \multicolumn{3}{c|}{\text{VOC, mAP$_{50-95}$}} & \multicolumn{3}{c}{\text{SKU-110k, mAP$_{50-95}$}}\\
\midrule
{Predictor metric} & {global $\tau$} & {top-15\% $\tau$} & {\%Pareto pred. (GPU)} & {global $\tau$} & {top-15\% $\tau$} & {\%Pareto pred. (GPU)}\\
\midrule    
{GraSP} & {-0.011} & {-0.068} & {0.062} & {0.040} & {0.032} & {0.025}\\
{Plain} & {0.029} & {0.069} & {0.015} & {-0.388} & {-0.176} & {0.025}\\
{JacobCov} & {0.095} & {-0.078} & {0.015} & {0.541} & {0.136} & {0.025}\\
{ZiCo} & {0.195} & {0.016} & {0.015} & {0.115} & {0.081} & {0.025}\\
{Zen} & {0.255} & {0.092} & {0.062} & {0.146} & {0.121} & {0.050}\\
{GradNorm} & {0.262} & {0.173} & {0.015} & {-0.331} & {-0.072} & {0.025}\\
{Fisher} & {0.280} & {0.156} & {0.015} & {-0.380} & {-0.096} & {0.025}\\
{L2 norm} & {0.326} & {0.090} & {0.015} & {0.189} & {0.118} & {0.025}\\
{SNIP} & {0.336} & {0.217} & {0.015} & {-0.290} & {-0.059} & {0.025}\\
{\#params} & {0.399} & {0.372} & {0.031} & {0.256} & {0.119} & {0.050}\\
{SynFlow} & {0.558} & {0.227} & {0.062} & {0.512} & {0.254} & {0.100}\\
{MACs} & {0.739} & {0.520} & {0.123} & {0.604} & {0.314} & {0.125}\\
{NWOT} & {0.756} & {0.622} & {0.262} & {0.703} & {0.321} & {\bf{0.200}}\\
{NWOT (pre-act)} & {{\bf 0.827}} & {{\bf 0.623}} & {{\bf 0.292}} & {{\bf 0.765}} & {{\bf 0.406}} & {{\bf 0.200}}\\
\midrule
{VOC training} & {0.847} & {0.665} & {0.369} & {0.739} & {0.374} & {0.425}\\
{from scratch (mAP$_{50-95}$)} &  {} & {} & {} & {}\\
\bottomrule
\end{tabularx}
\end{table*}

\begin{figure*}[ht!]
    \centering
    \includegraphics[width=1.0\textwidth]{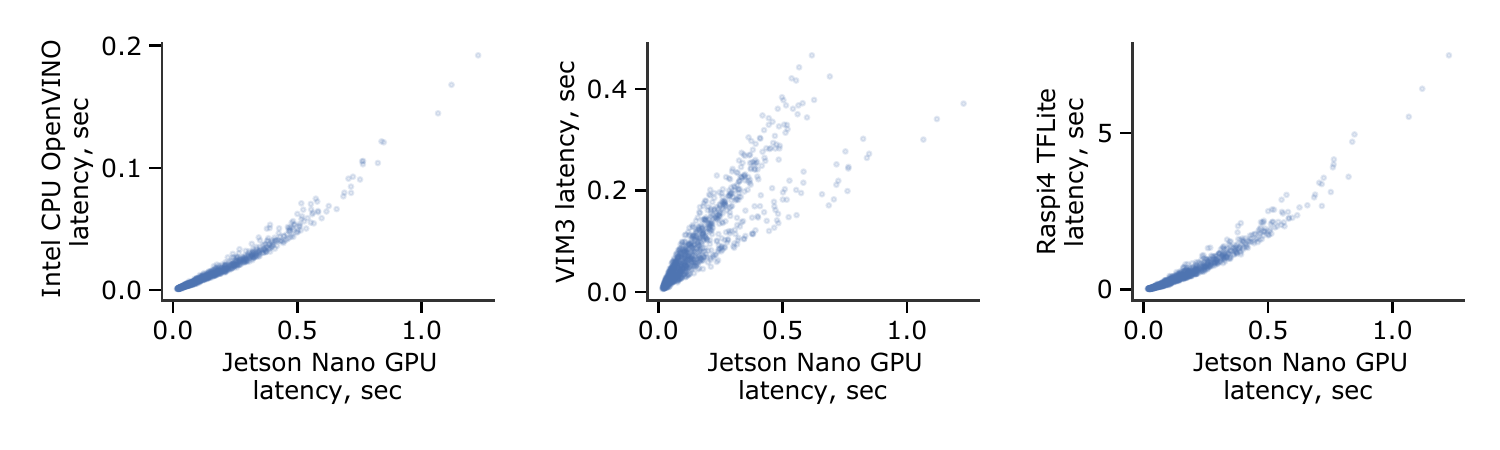}
    \vspace{2mm}
    \caption{Scatter plots of latency values measured for \textit{YOLOBench} models on the Jetson Nano GPU (ORT, FP32 precision) vs. latency values on other hardware platforms.}
    \label{fig:nano_timing_corr}
\end{figure*}

\begin{figure*}[ht!]
    \centering
    \includegraphics[width=1.0\textwidth]{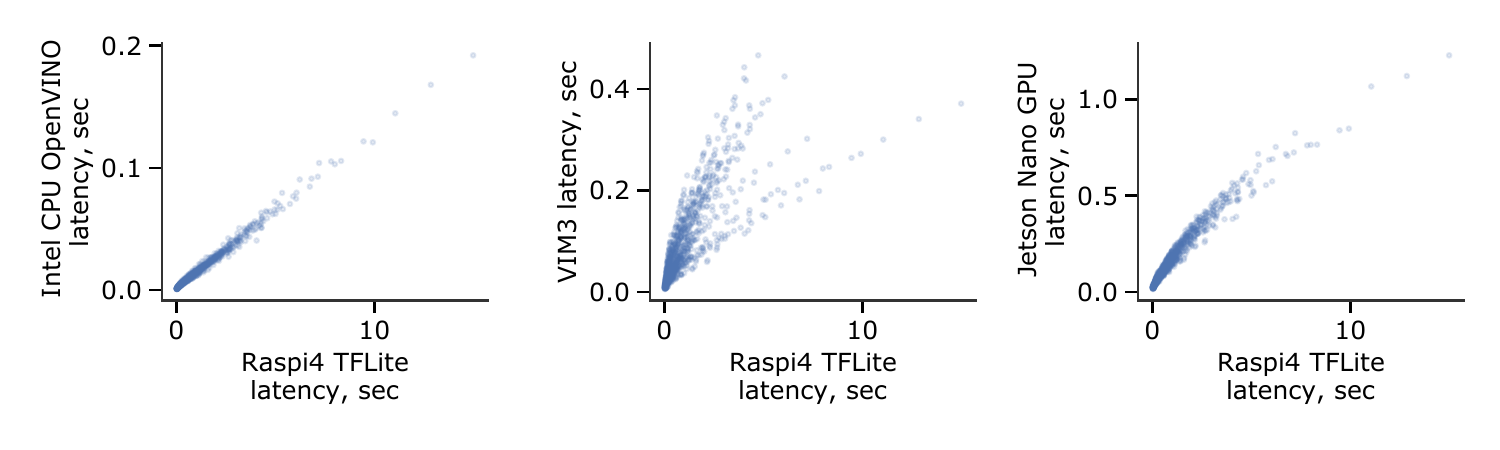}
    \vspace{2mm}
    \caption{Scatter plots of latency values measured for \textit{YOLOBench} models on the Raspberry Pi 4 CPU (TFLite with XNNPACK, FP32 precision) vs. latency values on other hardware platforms.}
    \label{fig:raspi4_timing_corr}
\end{figure*}

\begin{figure*}[ht!]
    \centering
    \includegraphics[width=0.6\textwidth]{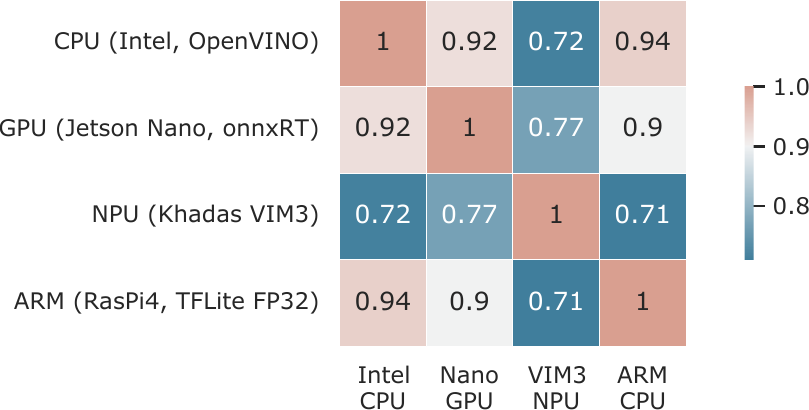}
    \vspace{2mm}
    \caption{Correlation matrix (Kendall-Tau scores are shown) for latency values on 4 hardware platforms/runtimes considered in \textit{YOLOBench}.}
    \label{fig:timing_corr_matrix}
\end{figure*}

\begin{figure*}[ht!]
    \centering
    \includegraphics[width=1.0\textwidth]{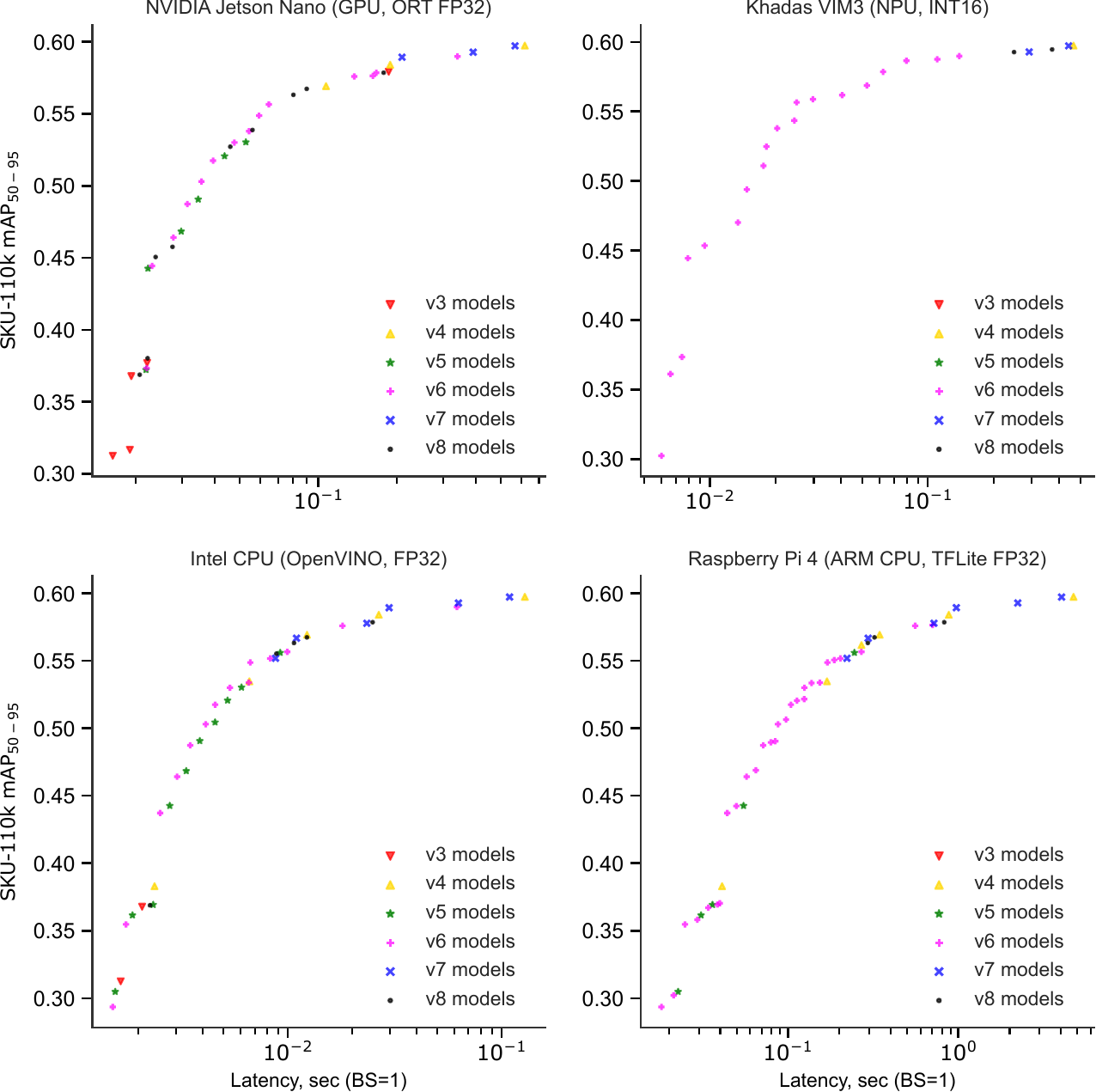}
    \vspace{2mm}
    \caption{Pareto frontiers of \textit{YOLOBench} models fine-tuned on the SKU-110k dataset (on several target resolutions) from COCO-pretrained weights on 4 different hardware platforms. Each point represents a single model in the mAP-latency space, with the model family coded with color and marker size (all YOLOv6-3.0 models are represented by the same color).}
    \label{fig:sku_pareto}
\end{figure*}

\begin{figure*}[ht!]
    \centering
    \includegraphics[width=1.0\textwidth]{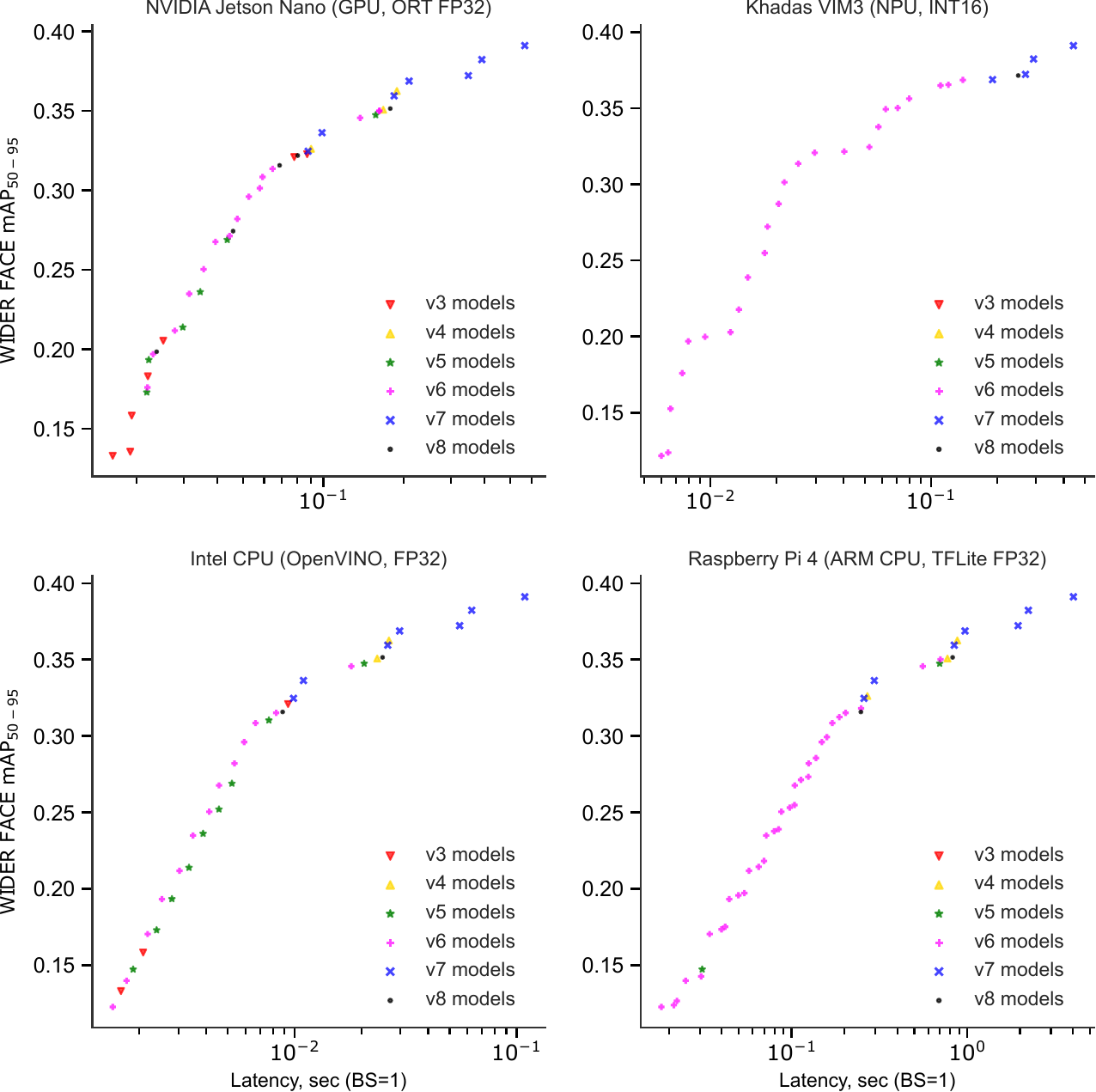}
    \vspace{2mm}
    \caption{Pareto frontiers of \textit{YOLOBench} models fine-tuned on the WIDER FACE dataset (on several target resolutions) from COCO-pretrained weights on 4 different hardware platforms. Each point represents a single model in the mAP-latency space, with the model family coded with color and marker size (all YOLOv6-3.0 models are represented by the same color).}
    \label{fig:wider_pareto}
\end{figure*}

\begin{figure*}[ht!]
    \centering
    \includegraphics[width=1.0\textwidth]{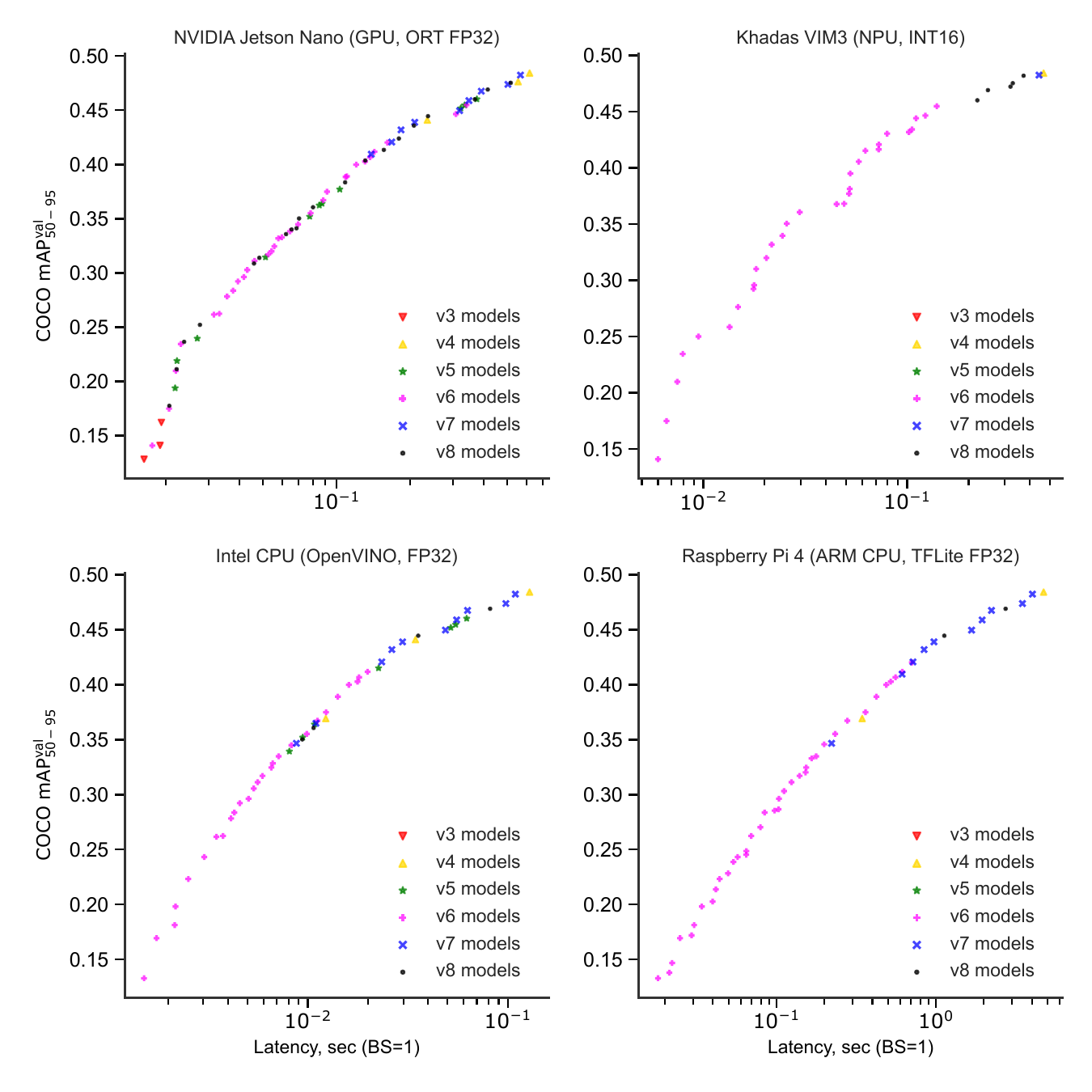}
    \vspace{2mm}
    \caption{Pareto frontiers of \textit{YOLOBench} models trained on the COCO dataset (640x640 image resolution) and validated on several target image resolutions on COCO minival (without additional fine-tuning) on 4 different hardware platforms. Each point represents a single model in the mAP-latency space, with the model family coded with color and marker size (all YOLOv6-3.0 models are represented by the same color).}
    \label{fig:coco_pareto}
\end{figure*}

\begin{figure*}[ht!]
    \centering
    \includegraphics[width=1.0\textwidth]{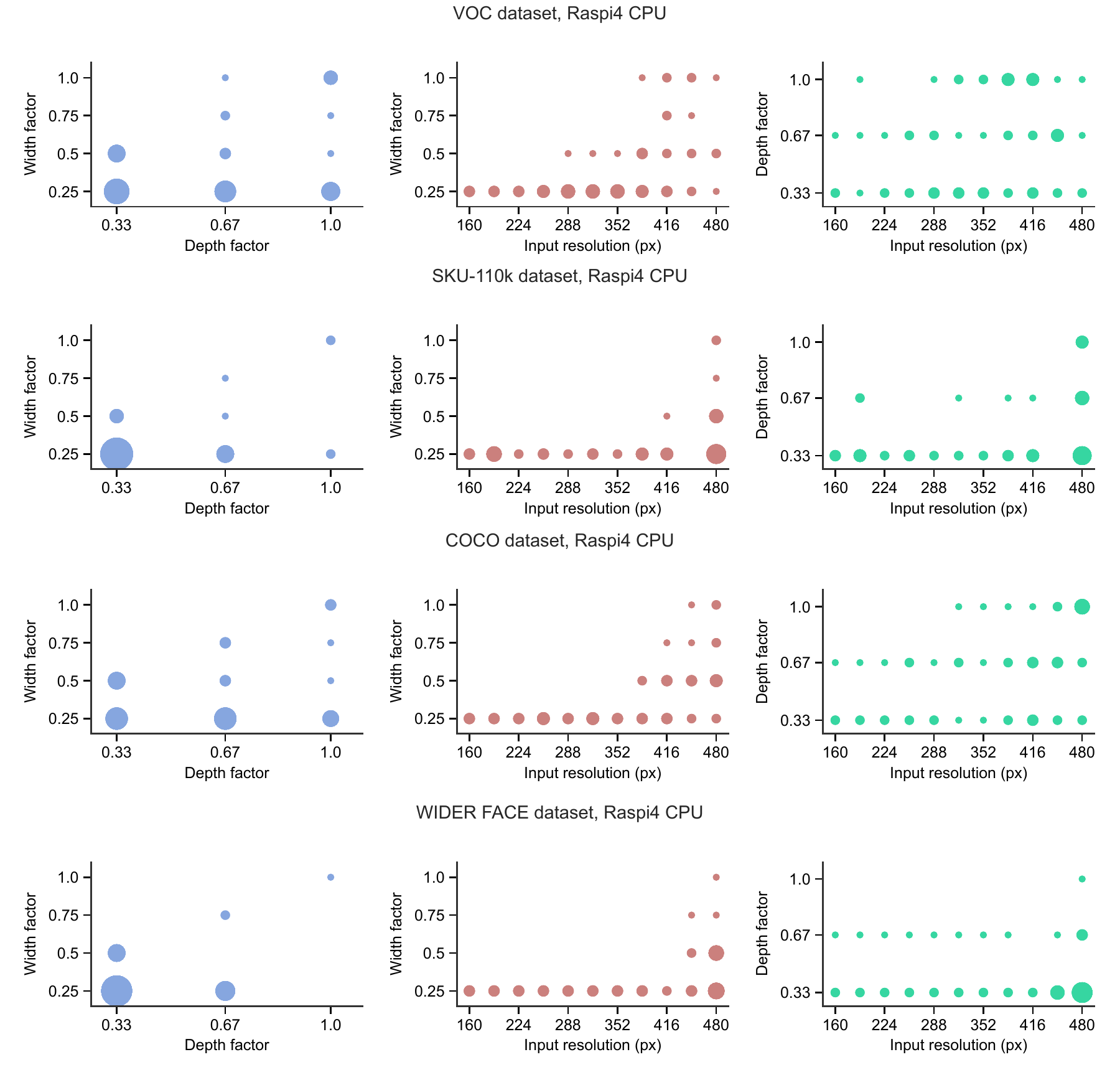}
    \vspace{2mm}
    \caption{Statistics of model scaling parameters (depth factor, width factor, input resolutions) in Pareto-optimal models on 4 different datasets with latency measured on the Raspberry Pi 4 ARM CPU (TFLite, FP32). The size of each point (circle) is proportional to the number of models for that parameter combination.}
    \label{fig:scaling_raspi4}
\end{figure*}

\begin{figure*}[ht!]
    \centering
    \includegraphics[width=1.0\textwidth]{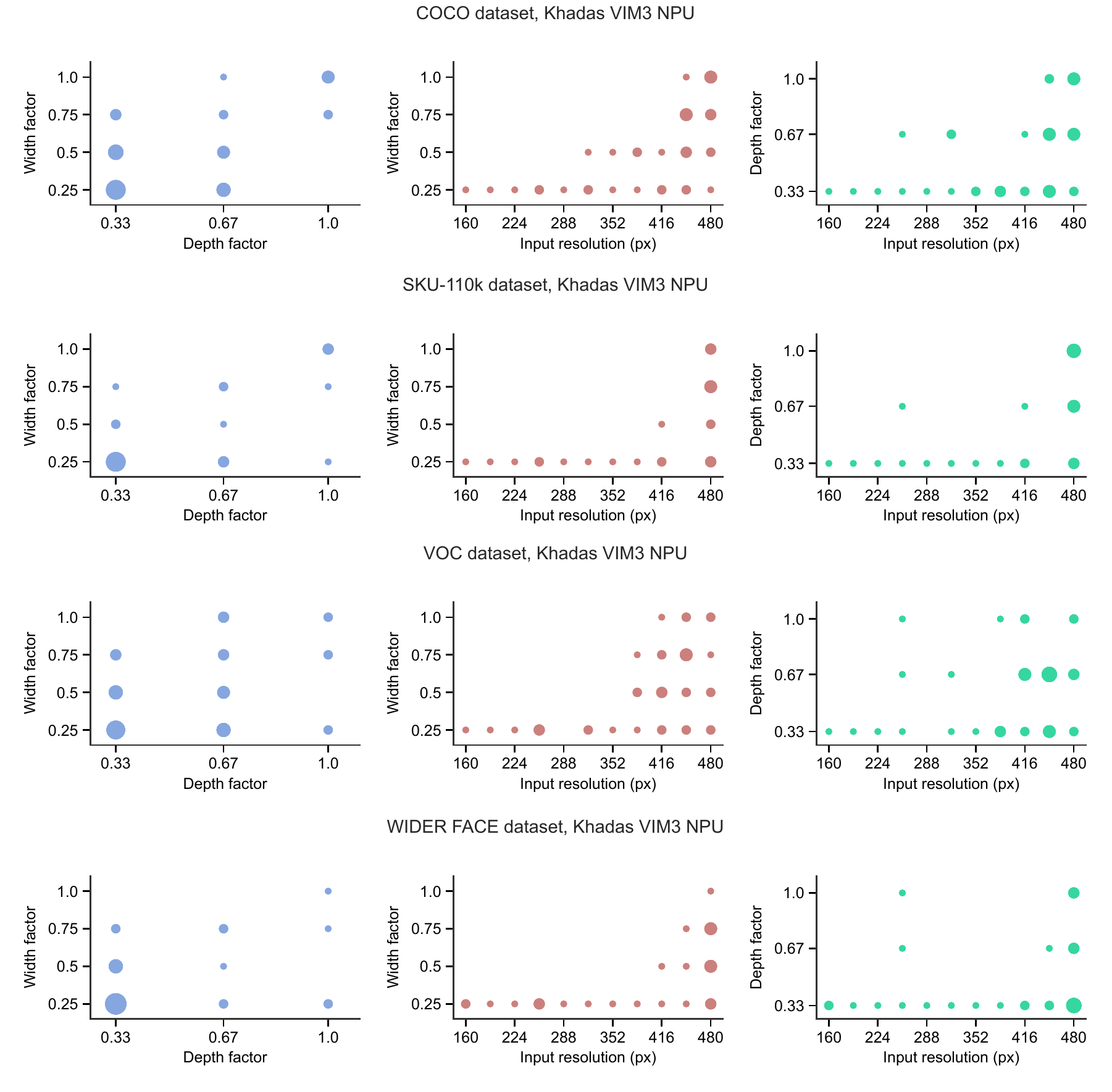}
    \vspace{2mm}
    \caption{Statistics of model scaling parameters (depth factor, width factor, input resolutions) in Pareto-optimal models on 4 different datasets with latency measured on Khadas VIM3 NPU (INT16). The size of each point (circle) is proportional to the number of models for that parameter combination.}
    \label{fig:scaling_vim3}
\end{figure*}

\begin{figure*}[ht!]
    \centering
    \includegraphics[width=1.0\textwidth]{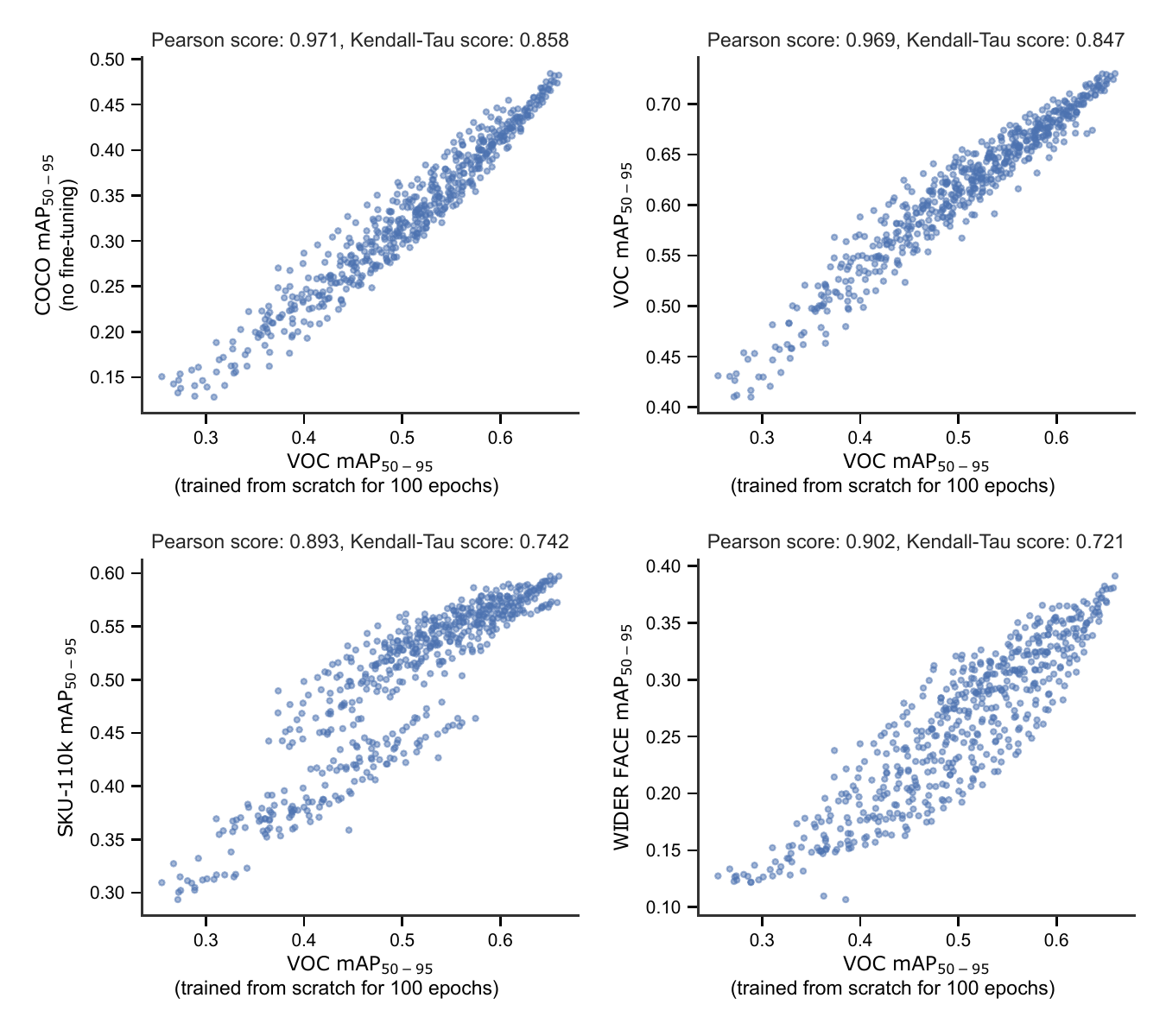}
    \vspace{2mm}
    \caption{Scatter plots of mAP$_{50-95}$ values obtained on 4 \textit{YOLOBench} datasets for models fine-tuned from COCO-pretrained weights (for all datasets except COCO) vs. mAP$_{50-95}$ of models trained on the VOC dataset from scratch for 100 epochs. High correlation of target metric and mAP of VOC training from scratch is observed for several datasets.}
    \label{fig:voc_scratch_corr}
\end{figure*}

\begin{figure*}[ht!]
    \centering
    \includegraphics[width=1.0\textwidth]{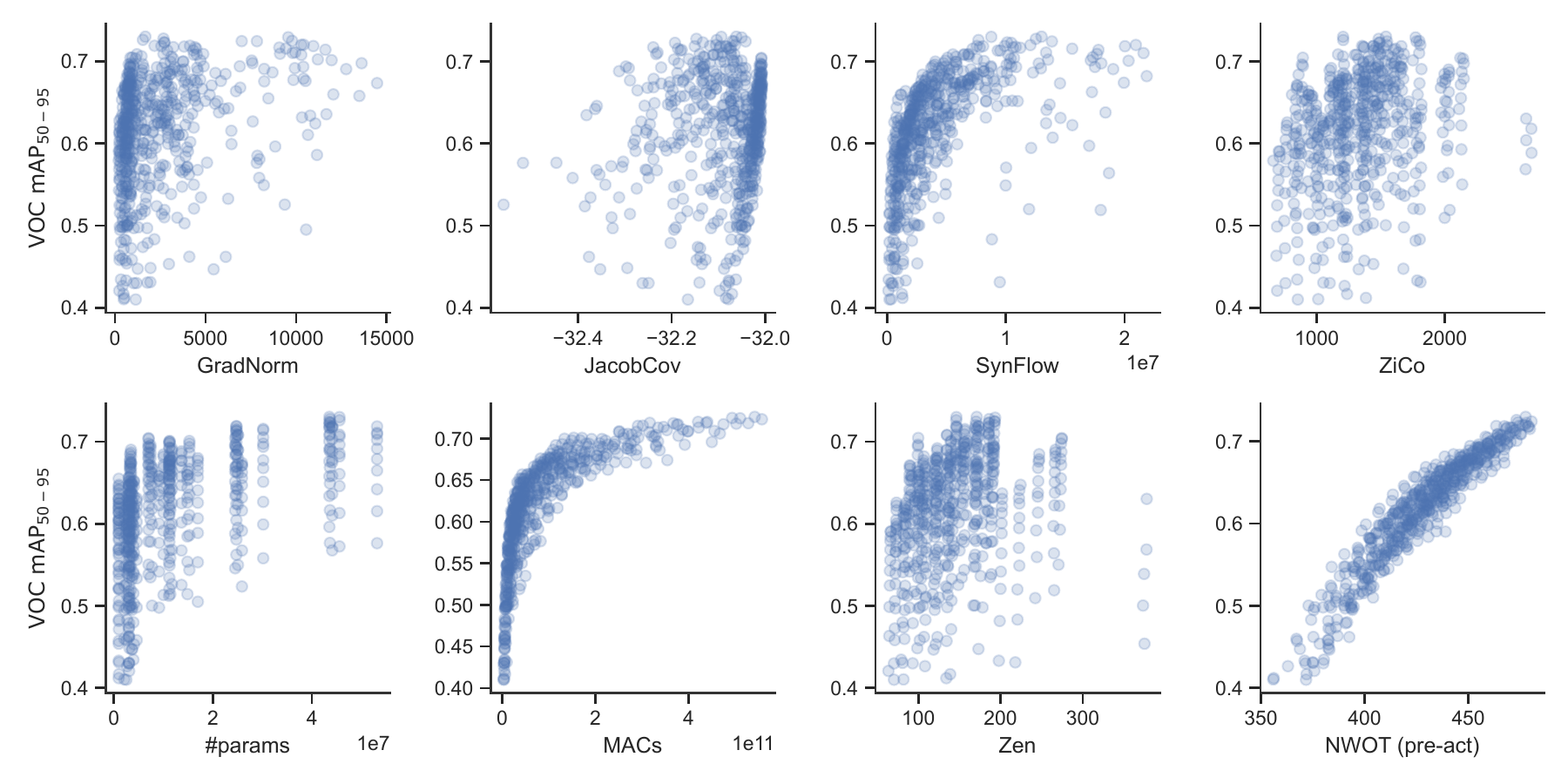}
    \vspace{2mm}
    \caption{Scatter plots of zero-cost predictor values computed on randomly initialized \textit{YOLOBench} models vs. mAP$_{50-95}$ of these models fine-tuned on VOC from COCO-pretrained weights.}
    \label{fig:ZC_scatter}
\end{figure*}

\begin{figure*}[ht!]
    \centering
    \includegraphics[width=0.7\textwidth]{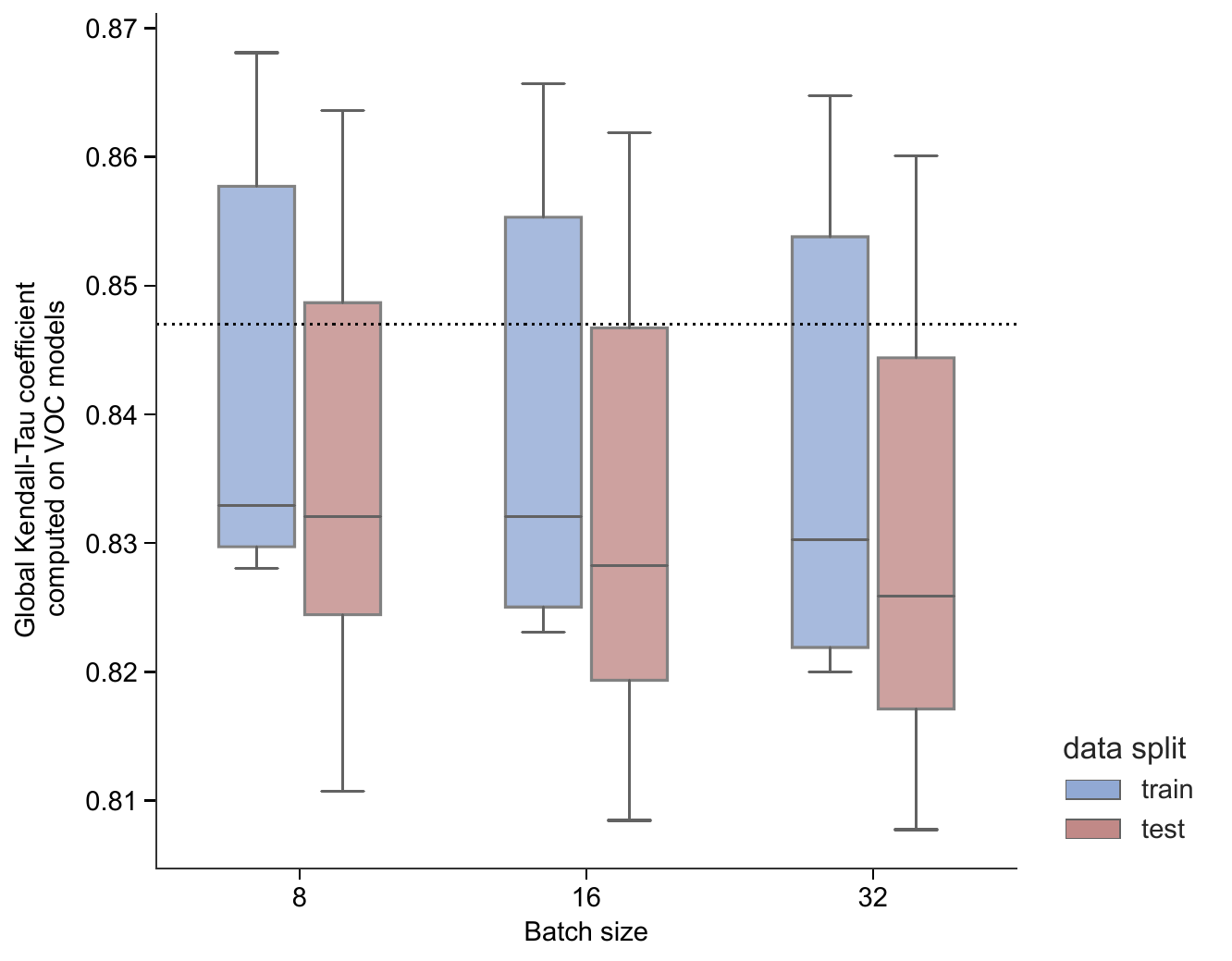}
    \vspace{2mm}
    \caption{Robustness of pre-activation NWOT estimator in predicting mAP$_{50-95}$ values of VOC models in \textit{YOLOBench}. Shown is Kendall-Tau score dependence on the batch size, as well as the data, split used to sample the batch (training (augmented) data vs. testing data (no augmentations)). The dotted vertical line corresponds to the performance of mAP$_{50-95}$ VOC in training from scratch used as a predictor.}
    \label{fig:nwot_robust}
\end{figure*}

\begin{table*}
  \small
  \begin{center}
    \caption{Mean and standard deviation of the global Kendall-Tau scores for NWOT metrics computed for 5 different randomly sampled batches of size 16 on VOC \textit{YOLOBench} models. The metric denoted as "no head" was computed only for the layers contained in the neck and backbone of YOLO models, not in the detection head. The second column shows Kendall-Tau scores for prediction with the mean ZC metric values averaged over the 5 batches.}
    \label{tab:nwot_robust_hohead}
    \vspace*{2mm}
    \begin{tabularx}{\linewidth}{X|X|X} 
      \hline
      {ZC metric} & {global $\tau$} & {global $\tau$ (prediction with mean ZC value)}\\
      \hline
      {NWOT} & {0.7839 (0.0159)} & {0.7895} \\
      \hline
      {NWOT (pre-act)} & {0.8402 (0.0191)} & {0.8486} \\
      \hline
      {NWOT (pre-act, no head)} & {\textbf{0.8472 (0.0194)}} & {\textbf{0.8570}} \\
      \hline
    \end{tabularx}
  \end{center}
\end{table*}

\begin{figure*}[ht!]
    \centering
    \includegraphics[width=0.85\textwidth]{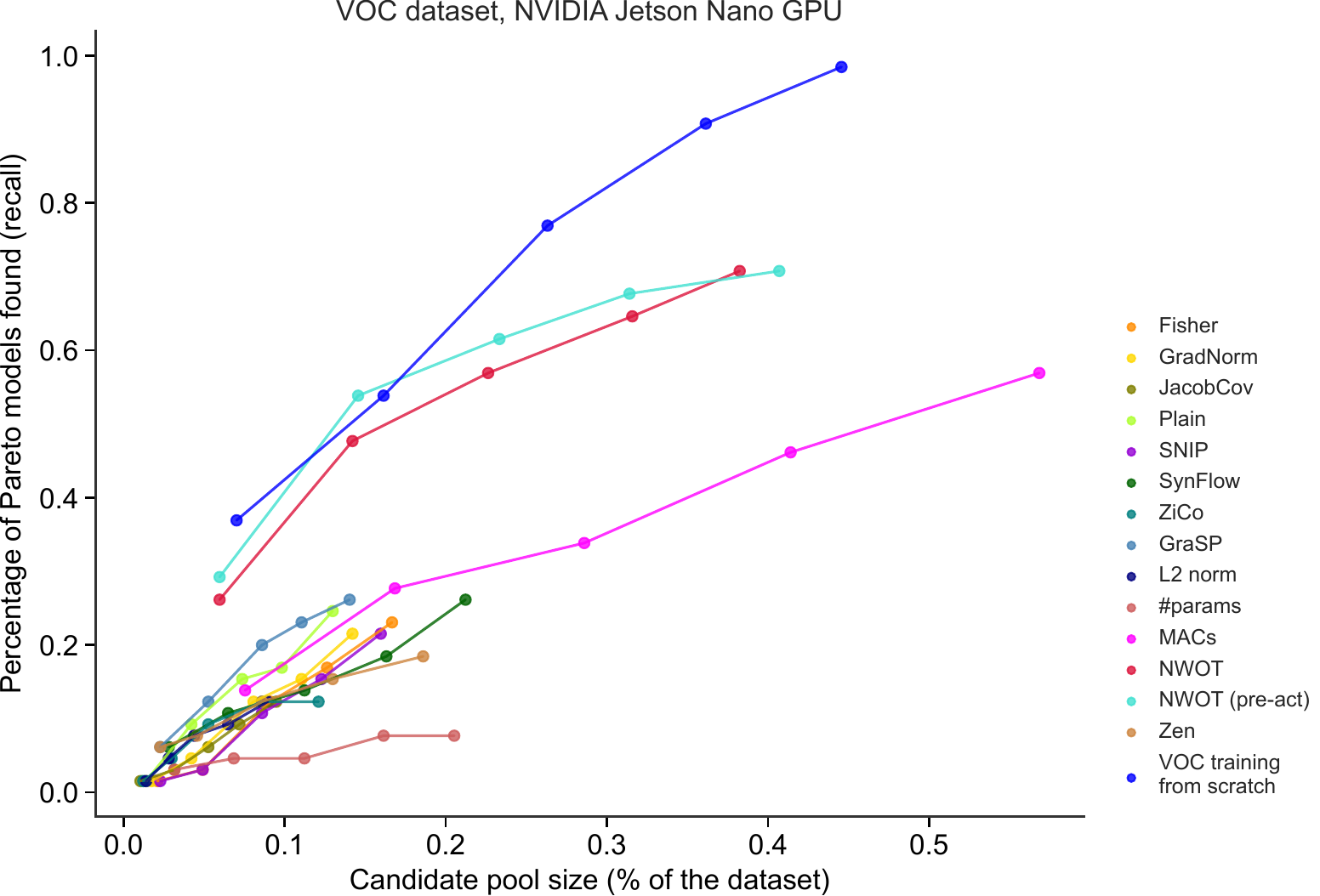}
    \vspace{2mm}
    \caption{Percentage of all actual Pareto models (recall) found in the candidate pools consisting of first $N$ ($N$ from 1 to 5) ZC-based Pareto sets on the VOC dataset with latency measured on the Jetson Nano GPU (ORT, FP32). Data shown for all zero-cost estimators considered in this study, in addition to mAP$_{50-95}$ values of VOC training from scratch used as a performance predictor.}
    \label{fig:ZC_pareto_VOC_nano}
\end{figure*}

\begin{figure*}[ht!]
    \centering
    \includegraphics[width=1.0\textwidth]{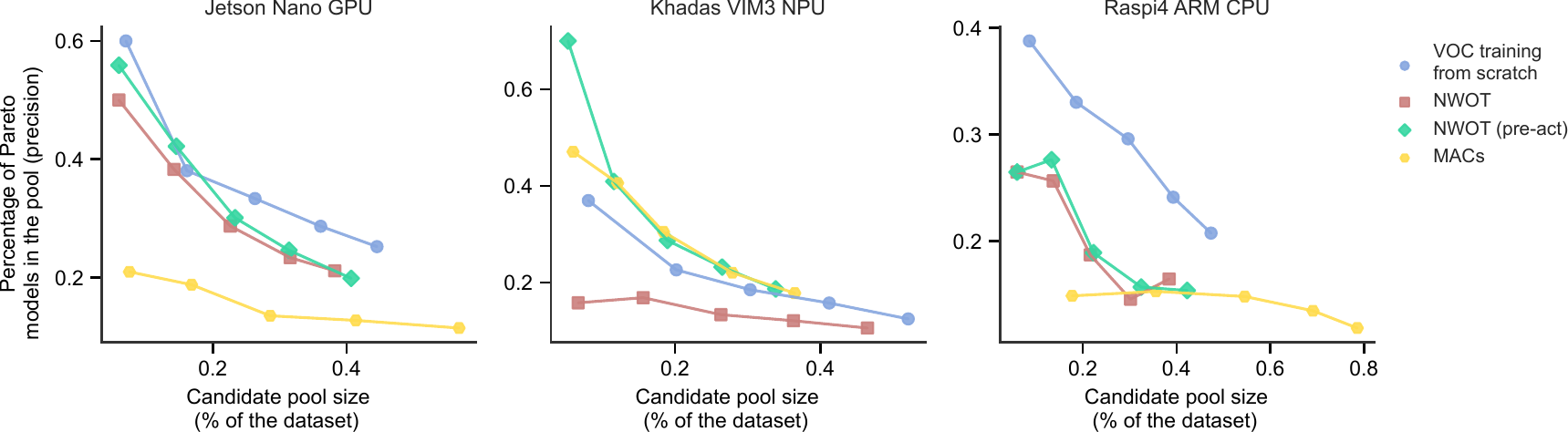}
    \vspace{2mm}
    \caption{Percentage of Pareto models out of all models found in the candidate pools (precision) consisting of first $N$ ($N$ from 1 to 5) ZC-based Pareto sets on the VOC dataset depending on the hardware platform and performance predictor used.}
    \label{fig:ZC_pareto_VOC_precision}
\end{figure*}

\begin{figure*}[ht!]
    \centering
    \includegraphics[width=1.0\textwidth]{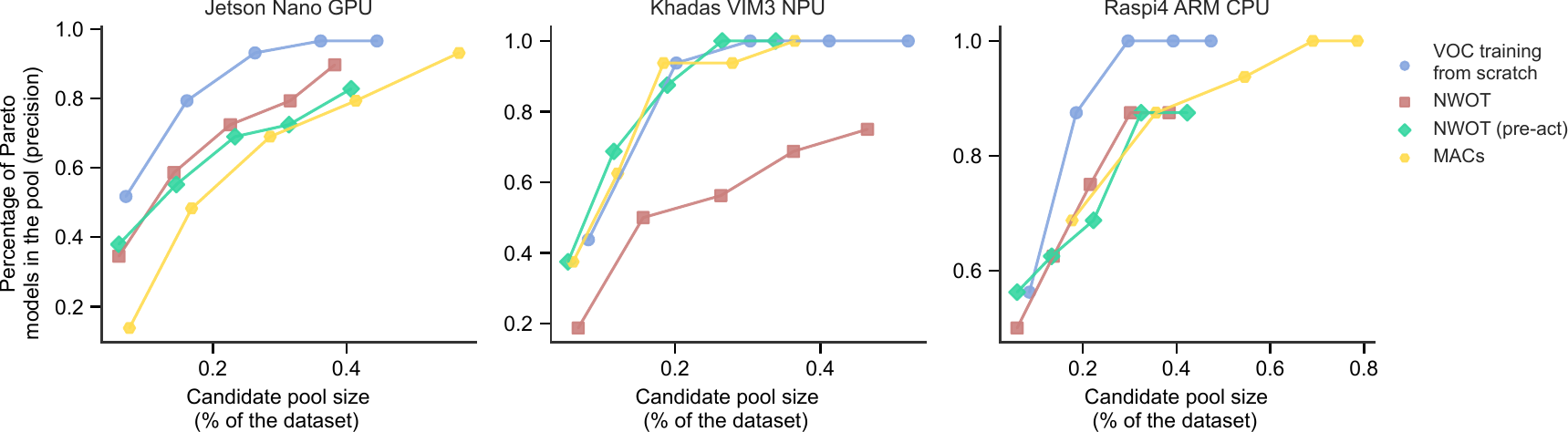}
    \vspace{2mm}
    \caption{Percentage of all actual Pareto models (recall) found in the candidate pools consisting of first $N$ ($N$ from 1 to 5) ZC-based Pareto sets on the VOC dataset depending on the hardware platform and performance predictor used. Models with different input resolutions but the same architecture as treated as a single model.}
    \label{fig:ZC_pareto_VOC_nores}
\end{figure*}

\begin{figure*}[ht!]
    \centering
    \includegraphics[width=1.0\textwidth]{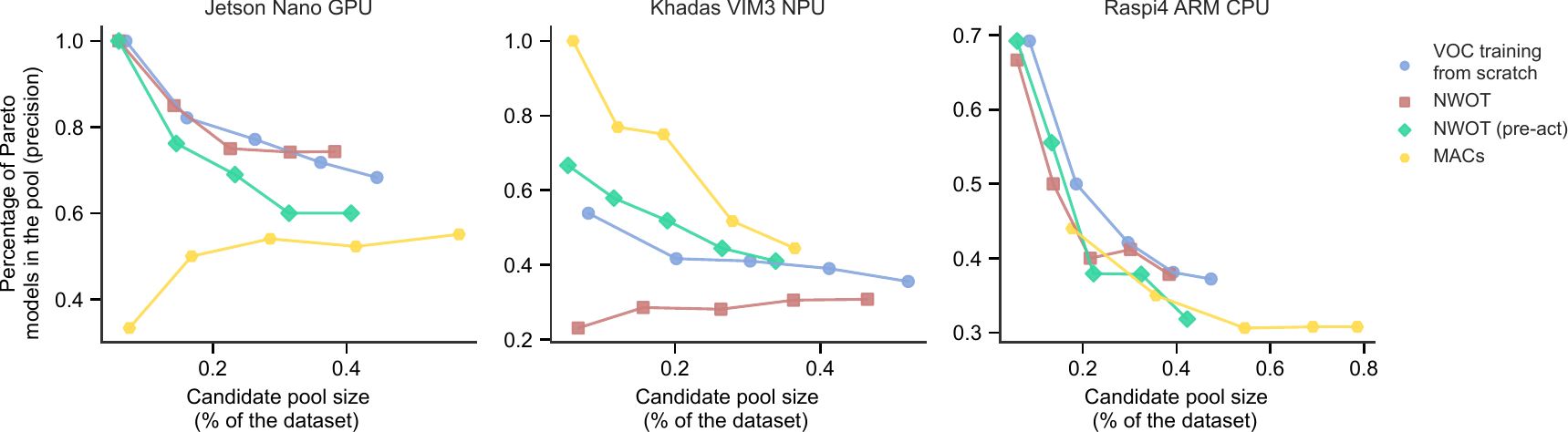}
    \vspace{2mm}
    \caption{Percentage of Pareto models out of all models found in the candidate pools (precision) consisting of first $N$ ($N$ from 1 to 5) ZC-based Pareto sets on the VOC dataset depending on the hardware platform and performance predictor used. Models with different input resolutions but the same architecture as treated as a single model.}
    \label{fig:ZC_pareto_VOC_nores_precision}
\end{figure*}

\begin{figure*}[ht!]
    \centering
    \includegraphics[width=1.0\textwidth]{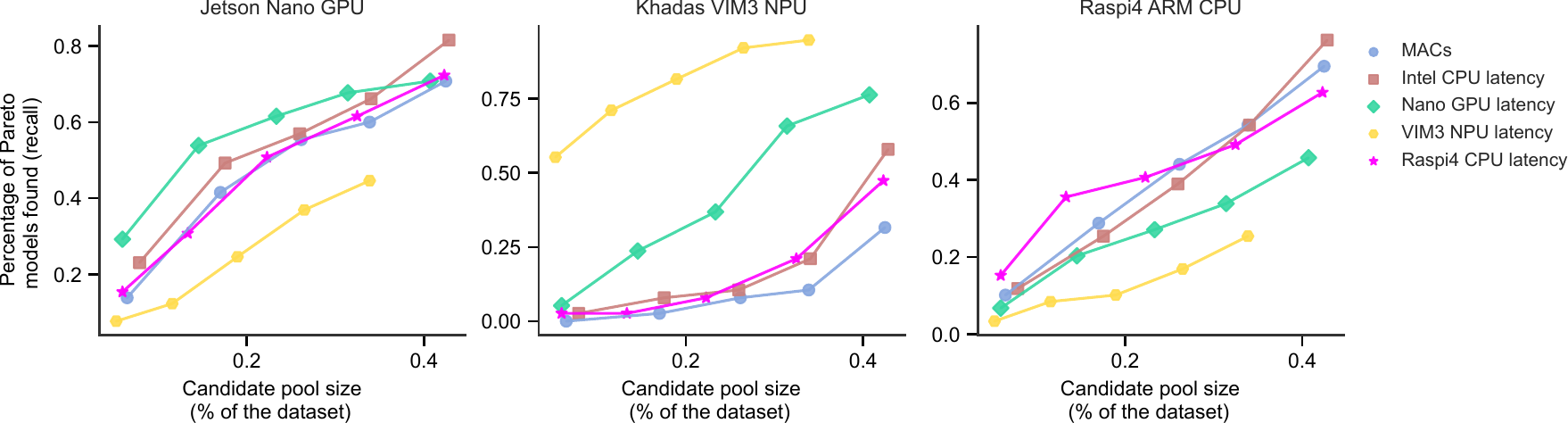}
    \vspace{2mm}
    \caption{Percentage of all actual Pareto models (recall) found in the candidate pools consisting of first $N$ ($N$ from 1 to 5) ZC-based Pareto sets on the VOC dataset with pre-activation NWOT score as accuracy estimator depending on the hardware platform and latency proxy used. Aside from the actual on-device latency, other latency proxies considered are MAC count and latency on other devices.}
    \label{fig:ZC_pareto_VOC_latproxy}
\end{figure*}

\begin{figure*}[ht!]
    \centering
    \includegraphics[width=1.0\textwidth]{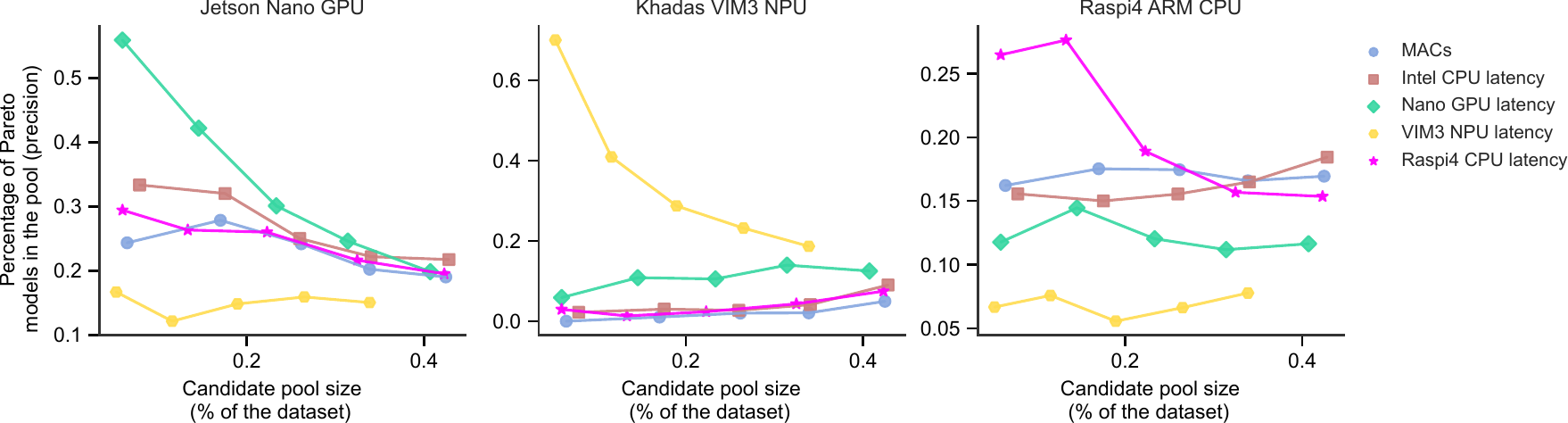}
    \vspace{2mm}
    \caption{Percentage of Pareto models out of all models found in the candidate pools (precision) consisting of first $N$ ($N$ from 1 to 5) ZC-based Pareto sets on the VOC dataset with pre-activation NWOT score as accuracy estimator depending on the hardware platform and latency proxy used. Aside from the actual on-device latency, other latency proxies considered are MAC count and latency on other devices.}
    \label{fig:ZC_pareto_VOC_latproxy_prec}
\end{figure*}

\begin{figure*}[ht!]
    \centering
    \includegraphics[width=1.0\textwidth]{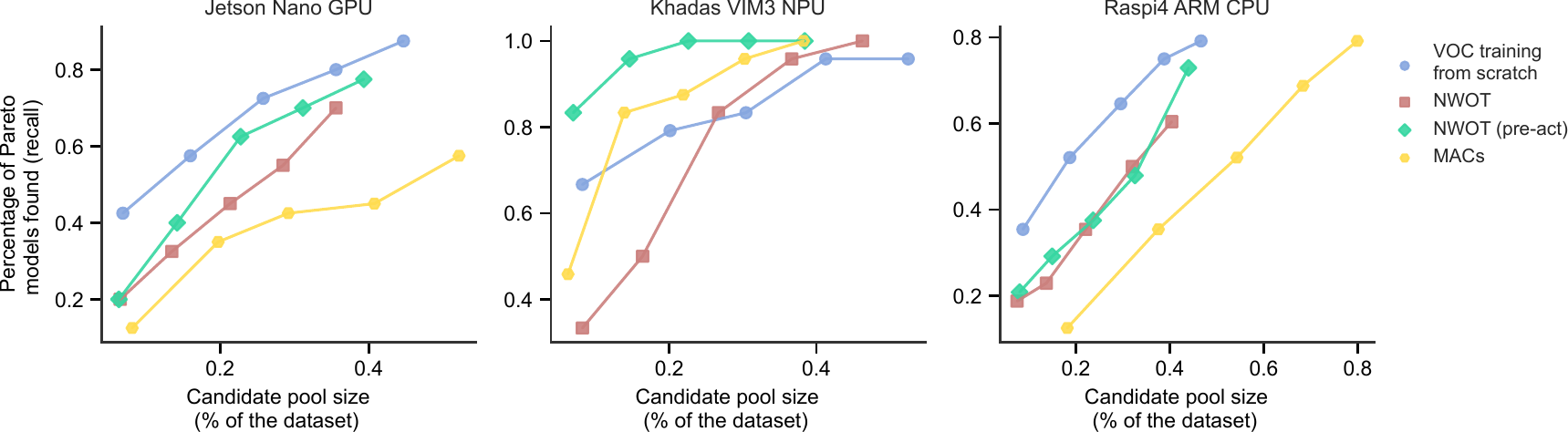}
    \vspace{2mm}
    \caption{Percentage of all actual Pareto models (recall) found in the candidate pools consisting of first $N$ ($N$ from 1 to 5) ZC-based Pareto sets on the SKU-110k dataset with latency measured on the Jetson Nano GPU (ORT, FP32). Data shown for MAC count and NWOT score as performance estimators, in addition to mAP$_{50-95}$ values of VOC training from scratch used as a performance predictor on the SKU-110k dataset.}
    \label{fig:zc_sku_parero_front}
\end{figure*}

\begin{table*}
\centering
\caption{Example YOLO-PAN-C3 models with \texttt{timm} backbones identified in the NWOT-latency Pareto frontier, with pre-activation NWOT score computed on the VOC dataset. Latency values are measured on Raspberry Pi 4 ARM CPU with TFLite (FP32), batch size 1.}
\label{tab:nwot_timm_pareto}
\begin{tabularx}{\textwidth}{l|X|X|X}
\toprule
Model name & Input resolution & Raspi4 CPU latency, sec & NWOT (pre-act) \\
\midrule
\texttt{yolo\_pan\_efficientnet\_b4} & { 480} & { 1.72} & { 511.84 }  \\
\texttt{yolo\_pan\_tf\_efficientnet\_b4\_ap} & { 480} & { 1.71} & { 511.77 }  \\
\texttt{yolo\_pan\_gc\_efficientnetv2\_rw\_t} & { 480} & { 1.41} & { 508.73 }  \\
\texttt{yolo\_pan\_tf\_efficientnet\_lite4} & { 480} & { 1.08} & { 506.67 }  \\
\texttt{yolo\_pan\_fbnetv3\_d} & { 480} & { 0.71} & { 502.71 }  \\
\texttt{yolo\_pan\_tf\_efficientnet\_lite1} & { 480} & { 0.61} & { 493.48 }  \\
\texttt{yolo\_pan\_efficientnet\_lite1} & { 480} & { 0.61} & { 493.32 }  \\
\texttt{yolo\_pan\_mobilenetv2\_110d} & { 480} & { 0.54} & { 480.92 }  \\
\texttt{yolo\_pan\_mobilenetv2\_075} & { 480} & { 0.45} & { 480.14 }  \\
\texttt{yolo\_pan\_tf\_mobilenetv3\_large\_075} & { 480} & { 0.45} & { 468.85 }  \\
\texttt{yolo\_pan\_mobilenetv2\_035} & { 480} & { 0.37} & { 457.41 }  \\
\texttt{yolo\_pan\_tf\_mobilenetv3\_small\_minimal\_100} & { 480} & { 0.36} & { 451.10 }  \\
\bottomrule
\end{tabularx}
\end{table*}

\begin{table*}
  \small
  \begin{center}
    \caption{COCO test mAP values and inference latency on Raspberry Pi 4 CPU (TFLite with XNNPACK backend, FP32) for YOLOv8s vs. a model identified from the NWOT-latency Pareto frontier (YOLO-FBNetV3-D-PAN). For mAP values, the mean and standard deviation over three random seeds are shown. For inference time, mean and standard deviation of inference time over 5
    runs (each one 100 iterations) are shown.} 
    \label{tab:timm_coco_test}
    \vspace*{2mm}
    \begin{tabularx}{\linewidth}{X|X|X|X|X|X|X|X} 
      \hline
      {Model} & {AP$^{test}_{50-95}$} & {AP$^{test}_{50}$} & {AP$^{test}_{75}$} & {AP$^{test}_{S}$} & AP$^{test}_{M}$ & AP$^{test}_{L}$ & {Latency, ms}\\
      \hline
      {YOLOv8s} & {43.17\% (0.12\%)} & {60.53\% (0.09\%)} & {46.5\% (0.08\%)} & {\textbf{22.7\%} (0.14\%)} & {47.13\% (0.17\%)} & {57.0\% (0.22\%)} & {1476.09 (1.49)} \\
      \hline
      {YOLOv8s-HSwish} & {42.90\% (0.0\%)} & {60.3\% (0.0\%)} & {46.30\% (0.0\%)} & {22.46\% (0.09\%)} & {47.0\% (0.08\%)} & {56.39\% (0.08\%)} & {1381.62 (7.34)} \\
      \hline
      {YOLO-FBNetV3-D-PAN} & {\textbf{43.87\%} (0.05\%)} & {\textbf{61.53\%} (0.09\%)} & {\textbf{47.23\%} (0.05\%)} & {22.67\% (0.19\%)} & {\textbf{47.87\%} (0.05\%)} & {\textbf{58.36\%} (0.12\%)} & {\textbf{1355.21} (9.93)} \\
      \hline
    \end{tabularx}
  \end{center}
\end{table*}

\begin{table*}
  \small
  \begin{center}
    \caption{COCO minival mAP and inference latency on Raspberry Pi 4 CPU (TFLite with XNNPACK backend, FP32) for YOLOv8s vs. a model identified from the NWOT-latency Pareto frontier (YOLO-FBNetV3-D-PAN). Mean and standard deviation of inference time over 5 runs (each one 100 iterations) are shown. The input image resolution used was 640x640, batch size $=1$ for latency measurements. Models were trained for 300 epochs, with batch size $=64$.} 
    \label{tab:timm_coco_full}
    \vspace*{2mm}
    \begin{tabularx}{\linewidth}{X|X|X} 
      \hline
      {Model} & {COCO mAP$^{val}_{50-95}$} & {Raspberry Pi 4 ARM CPU latency, ms}\\
      \hline
      {YOLOv8s} & {43.64\%} & {1476.09 (1.49)} \\
      \hline
      {YOLOv8s-HSwish} & {43.55\%} & {1381.62 (7.34)} \\
      \hline
      {YOLO-FBNetV3-D-PAN} & {44.63\%} & {1355.21 (9.93)} \\
      \hline
      {YOLO-FBNetV3-D-PAN-ReLU} & {44.07\%} & {1344.50 (8.06)} \\
      \hline
    \end{tabularx}
  \end{center}
\end{table*} \fi

\end{document}